%% file: main.tex
\documentclass[10pt,twocolumn,letterpaper]{article}
\usepackage[T1]{fontenc}
\usepackage[pagenumbers]{cvpr}
\usepackage{colortbl}
\usepackage{placeins}
\usepackage{float}
\usepackage[most]{tcolorbox}
\usepackage{listings}
\definecolor{revimshade}{RGB}{237,240,244}
\definecolor{revimourshade}{RGB}{226,241,237}
\definecolor{revimcA}{RGB}{36,91,166}
\definecolor{revimcB}{RGB}{30,117,164}
\definecolor{revimcC}{RGB}{28,141,147}
\definecolor{revimcD}{RGB}{46,156,116}
\definecolor{revimcE}{RGB}{83,169,83}
\newcommand{\revimname}{\textbf{\textcolor{revimcA}{R}\textcolor{revimcB}{e}\textcolor{revimcC}{V}\textcolor{revimcD}{i}\textcolor{revimcE}{M}}}

\definecolor{revimlink}{rgb}{0.21,0.49,0.74}
\definecolor{revimteal}{RGB}{0,112,125}
\definecolor{revimgreen}{RGB}{0,125,65}
\usepackage[breaklinks,colorlinks,allcolors=revimlink]{hyperref}

\title{ReViM: Rethinking the Memory Lifecycle through Associative Memory and Experience Learning for Agentic Long Video Understanding}
\author{
    Junbo Wang, Yuhao Qin, Yuke Li, Yining Zhu\\[3pt]
    Northwestern Polytechnical University, Xi'an, China\\[2pt]
    {\small\texttt{jbwang@nwpu.edu.cn, qinyuhaoswu@163.com}}\\
    {\small\texttt{liyuke@nwpu.edu.cn, yiningzhu@nwpu.edu.cn}}
}
\hypersetup{pdfauthor={Junbo Wang, Yuhao Qin, Yuke Li, Yining Zhu}}

\begin{document}
\maketitle
\raggedbottom
\begin{abstract}
Long video understanding requires reasoning over temporally dispersed evidence under the context constraints of multimodal large language models. However, prebuilt memories commit processing and abstraction before downstream needs are known, which can leave unused content and missing details. Question-centered agents can revisit relevant scenes yet repeat unproductive searches, obtain insufficient information, and leave acquired knowledge and retrieval adjustments confined to individual tasks. We introduce ReViM, a training-free framework connecting memory construction, refinement, and experience learning through a continuous memory lifecycle. Specifically, (i) associative episodic memory links observations through shared relation schemas while preserving occurrence-specific role bindings, enabling contextual retrieval through relative activation; (ii) retrieval refinement translates unmet information needs into focused acquisition requests using current feedback and applicable experience; and (iii) joint experience learning generalizes the goals, conditions, completion checks, and actions of executed adjustments together, preserving their shared arguments for reuse across videos. Newly acquired information enriches video memory, while acquisition outcomes provide feedback for experience learning without reference answers. Video knowledge thus accumulates across questions, and retrieval procedures transfer across videos. Experiments on four benchmarks demonstrate consistent improvements across three backbones. With GPT-5, ReViM outperforms the compared methods and improves accuracy over direct inference by 14.3 percentage points on LVBench and 12.0 points on LongVideoBench Long.
\end{abstract}
\section{Introduction}
\label{sec:introduction}

\begin{figure}[t]
    \centering
    \includegraphics[width=\columnwidth]{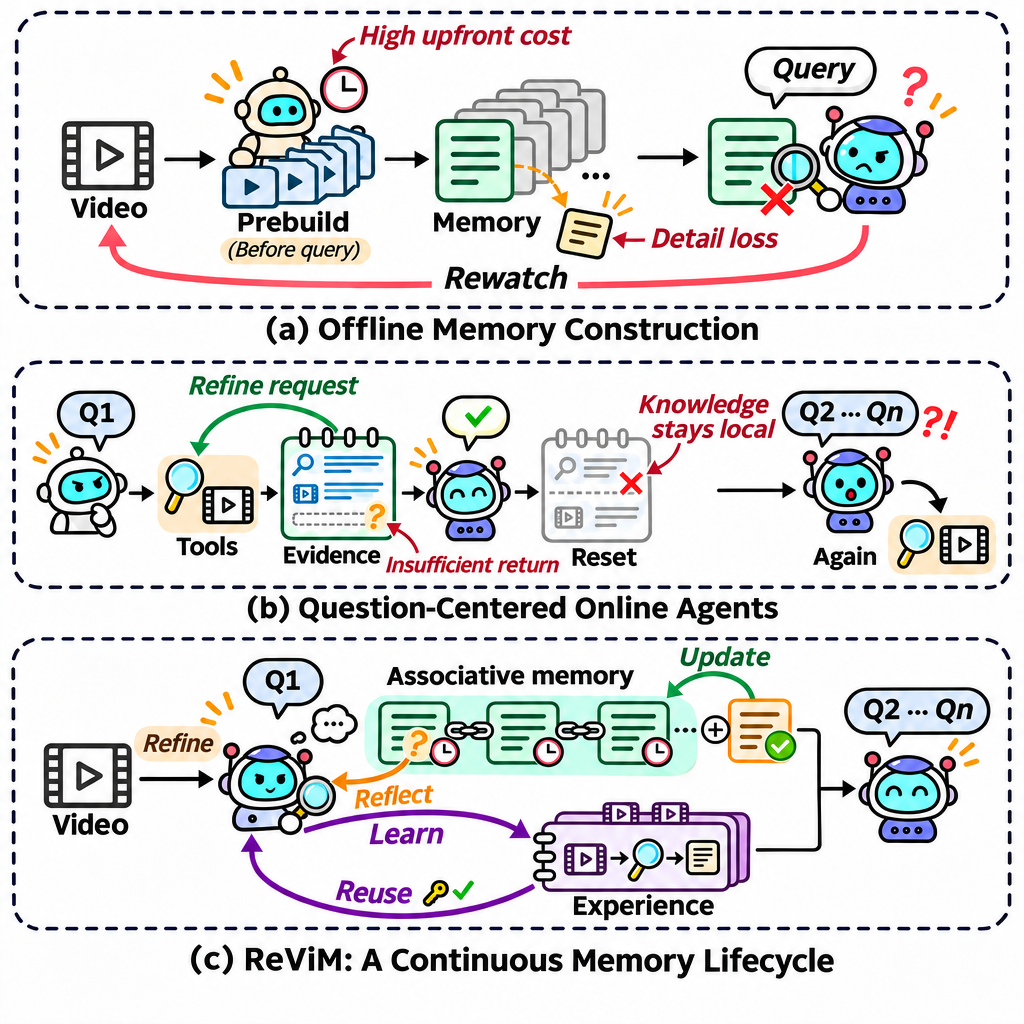}
    \caption{\textbf{Comparison of memory workflows for long video understanding.}
    (a) Offline construction incurs upfront processing and can omit details that require revisiting the video.
    (b) Question-centered agents refine insufficient returns, but question-local evidence must be reacquired for later questions.
    (c) ReViM updates associative memory across questions and learns acquisition procedures for reuse across videos.}
    \label{fig:motivation}
\end{figure}

Long video understanding requires recognizing brief actions, tracking entities, and relating distant events in extended recordings~\cite{wang2024videoagent,song2024moviechat}. Multimodal large language models (MLLMs)~\cite{liu2023visual,bai2025qwen25vltechnicalreport,chen2024expanding} and video instruction tuning~\cite{lin2024video,cheng2024videollama,zhang2024llava} have advanced temporal understanding. Despite richer video inputs~\cite{bai2025qwen3,hong2025glm}, sparse decisive events and redundant observations still strain context and computation budgets.

Longer visual contexts~\cite{zhang2024long,chen2025longvila}, compression~\cite{shen2024longvu,li2026videochat,wang2024retake,wang2025adaretake}, frame selection~\cite{tang2025adaptive}, and reasoning over local captions~\cite{zhang2024simple} improve limited-input processing. External memory retains processed content for retrieval without placing the entire video in context~\cite{luo2026video,jeong2025videorag,xue2025adavideorag}. Its construction, use, and improvement, however, are often handled separately.

Structured memory organizes episodes, entities, and event relations independently of the question~\cite{long2026seeing,yeo2026worldmm,choi2026keep,chen2026memdreamer}. It provides global context and amortizes construction across multiple questions, but commits processing effort and abstraction granularity before specific needs are known. With few questions, processing unrelated regions creates unused cost; selected summaries can also omit brief attributes, actions or relations needed later. Retrieval can recover represented information, whereas omitted details require returning to the source (Fig.~\ref{fig:motivation}(a)). Memory should therefore grow with actual demand and incorporate new observations into its existing organization.

Active agents select observations from the question and accumulated information~\cite{wang2024videoagent,li2026lenswalk,lin2026videoseek}. Their tools and working memories support exploration and focused inspection~\cite{zhang2026deep,yin2026videoarm,wang2026routing}, with feedback directing searches toward missing information. However, queries can be overly broad or poorly localized, and successive calls can return the same material; relevant returns may still leave an essential relation or attribute unresolved. Effective feedback must therefore identify the missing information and guide a concrete acquisition adjustment. Across questions, acquired observations and established relations often remain in the current execution, so later questions must reacquire context or reconstruct earlier understanding (Fig.~\ref{fig:motivation}(b)). 

Execution yields experience about unmet needs, acquisition adjustments and their outcomes. When confined to task-local trajectories, these connections cannot improve later acquisition. ExpeL and Agent Workflow Memory retain insights and workflows from experience~\cite{zhao2024expel,wang2024agent}. For video understanding, such procedures must connect their applicability conditions to executed requests and adapt to new entities and temporal contexts. Inspired by how people retain knowledge and improve actions through experience, we study a \emph{memory lifecycle} preserving both \emph{what has been understood} and \emph{how missing information was acquired}.

We introduce \textbf{ReViM}, a training-free framework with three connected components (Fig.~\ref{fig:motivation}(c)). \emph{Associative episodic memory} links contextual episodes through shared schemas and occurrence bindings, retrieving content through relative activation. \emph{Retrieval refinement} uses current feedback and instantiated experience to acquire missing information. \emph{Joint experience learning} generalizes goals, conditions, completion checks and requests with shared arguments, using acquisition outcomes to maintain experience. New observations enrich persistent knowledge.

\noindent Our contributions are threefold:
\begin{itemize}[topsep=0pt,itemsep=0pt,parsep=0pt,partopsep=0pt]
    \item We formulate a memory lifecycle connecting knowledge formation on demand, refinement during use, and retrieval experience across videos.
    \item We introduce a memory graph associating episodes through shared schemas and occurrence bindings, with retrieval based on relative activation.
    \item We develop joint experience learning that preserves dependencies among goals, conditions, completion requirements, and actions for acquisition in new videos.
\end{itemize}
\section{Related Work}
\label{sec:related-work}

\subsection{Long Video Understanding with MLLMs}

Visual instruction tuning connects language reasoning with perception~\cite{liu2023visual}. Video-LLaVA and VideoLLaMA~2 extend learning to video~\cite{lin2024video,cheng2024videollama}, while LLaVA-Video develops video instruction data~\cite{zhang2024llava}. Recent backbones further improve video understanding~\cite{bai2025qwen25vltechnicalreport,chen2024expanding,guo2025seed1,hong2025glm}. LongVA and LongVILA extend visual context~\cite{zhang2024long,chen2025longvila}; LongVU and VideoChat-Flash compress representations~\cite{shen2024longvu,li2026videochat}; ReTaKe and AdaReTaKe combine frame selection with cache compression~\cite{wang2024retake,wang2025adaretake}; and AKS balances relevance with temporal coverage~\cite{tang2025adaptive}. ReViM studies the organization and reuse of acquired understanding.

\subsection{Agentic Long Video Understanding}

LLoVi reasons over captions~\cite{zhang2024simple}, while tool frameworks~\cite{yang2023mm,suris2023vipergpt} and ReAct~\cite{yao2022react} support interactive reasoning. VideoAgent acquires information iteratively~\cite{wang2024videoagent}; VideoTree builds query-adapted representations~\cite{wang2025videotree}; DrVideo augments a video document~\cite{ma2025drvideo}; and DVD searches a prebuilt database~\cite{zhang2026deep}. AVP, LensWalk, and VideoSeek schedule observations~\cite{wang2026active,li2026lenswalk,lin2026videoseek}. Route2Look routes acquisition among browsing, grounding, and semantic retrieval~\cite{wang2026routing}, while MR.~Video applies MapReduce to captioning and question-conditioned analysis~\cite{pang2025mrvideo}. ReViM preserves knowledge across questions and learns procedures from retrieval outcomes.

\subsection{Video Memory and Retrieval}

MovieChat and MA-LMM maintain compact memories~\cite{song2024moviechat,he2024ma}. Retrieval-augmented generation~\cite{lewis2020retrieval} supports auxiliary text in Video-RAG~\cite{luo2026video}, video-corpus retrieval in VideoRAG~\cite{jeong2025videorag}, adaptive retrieval in AdaVideoRAG~\cite{xue2025adavideorag}, and requested-detail indexing in iRAG~\cite{arefeen2024irag}.

M$^3$-Agent builds entity-centered episodic and semantic memory~\cite{long2026seeing}; WorldMM combines modalities and timescales~\cite{yeo2026worldmm}; MERIT uses episodic keys and temporal expansion~\cite{choi2026keep}; MemDreamer organizes hierarchical graph memory~\cite{chen2026memdreamer}; and MM-Mem learns memory compression~\cite{lian2026verbatim}. VideoLucy adds detail through backtracking~\cite{zuo2026videolucy}, while VideoARM organizes sensory inputs, tool results, and working state~\cite{yin2026videoarm}. Sprout grows a temporal memory tree during answering~\cite{chen2026sproutbuildingdynamicmemory}; VideoTapestry refines a prebuilt hierarchy per query~\cite{liu2026videotapestryqueryadaptivememoryrefinement}.

\begin{figure*}[t]
    \centering
    \includegraphics[width=\textwidth]{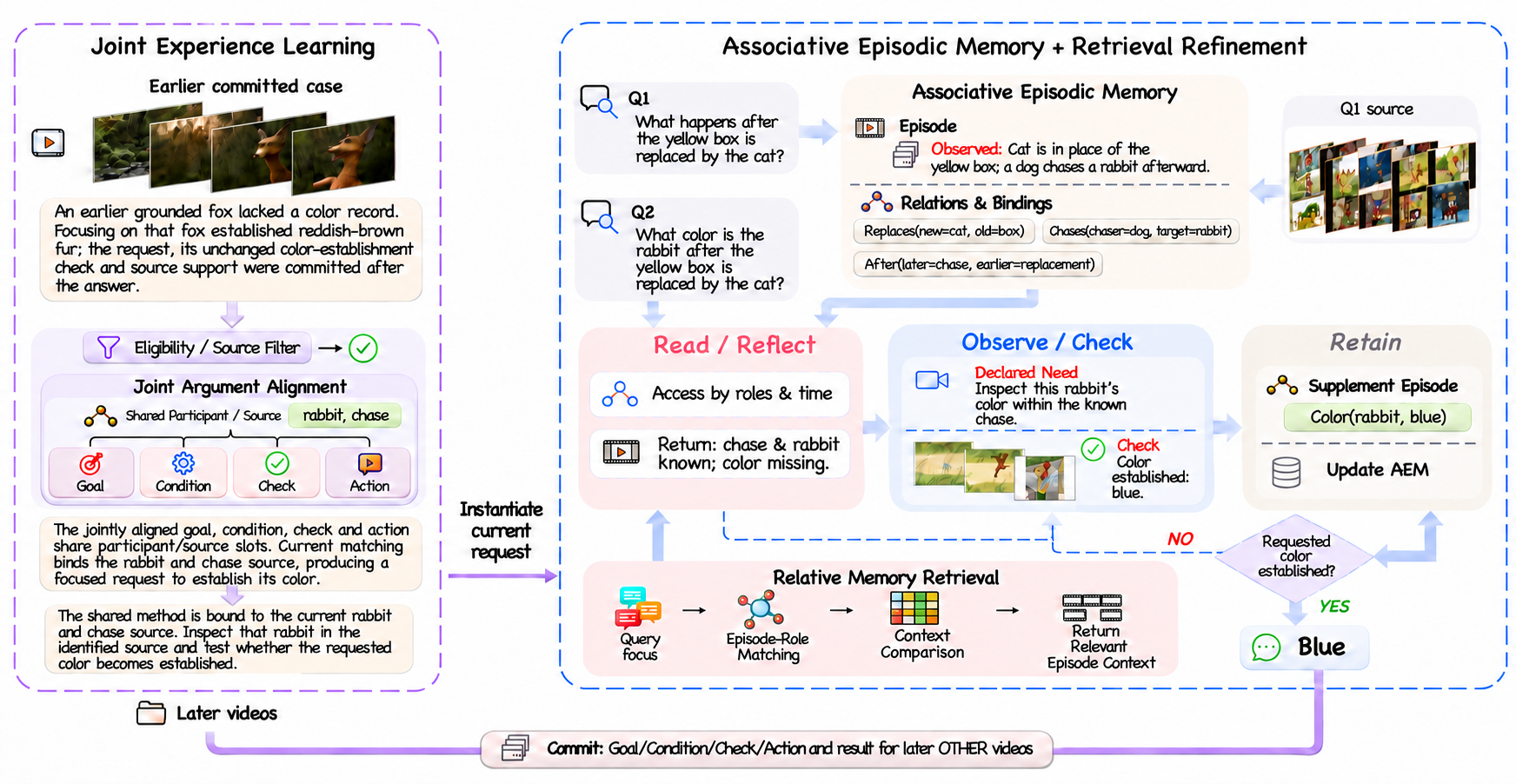}
    \caption{\textbf{ReViM framework with an illustrative execution.}
    Left: joint experience learning generalizes linked goals, conditions, checks and actions for the current video.
    Right: associative memory retains episodes; retrieval refinement acquires a missing attribute, checks the return and updates memory. New cases are committed after prediction for other videos.}
    \label{fig:framework}
\end{figure*}

\section{Method}\label{sec:revim-method}

\subsection{Overview}\label{sec:revim-overview}

ReViM maintains persistent video knowledge and transferable retrieval experience through three connected components (Fig.~\ref{fig:framework}): (i) \textbf{associative episodic memory} organizes observations for reuse across questions; (ii) \textbf{retrieval refinement} adjusts information acquisition using current feedback and applicable experience; and (iii) \textbf{joint experience learning (JEL)} turns executed adjustments into methods for subsequent videos.

Given a video \(V\) of duration \(T\), questions \(q_1,\ldots,q_K\) arrive sequentially. Video memory starts empty and persists across questions about \(V\). Experience memory starts empty for each independent evaluation stream; a question accesses only previously committed cases, excluding those from its own video before generalization.

The agent interleaves reasoning with tool execution~\cite{yao2022react}. At acquisition step \(t\), \textbf{Search} retrieves video records, \textbf{Read} expands a returned episode or revisits its media, and \textbf{Observe} samples a selected interval. Returns update the question's evidence \(E_t\) and request history \(H_t\); new observations enter video memory. Unmet needs invoke retrieval refinement. The model performs perception, semantic interpretation and reasoning, while the program performs graph retrieval, condition evaluation and experience generalization. Execution stops with an answer or after \(T_{\max}\) acquisition actions.

Each question rebuilds its evidence and history from material actually read. An empty memory starts with observation; retained episodes support an answer or guide focused inspection.

\subsection{Associative Episodic Memory}\label{sec:revim-associative-memory}

\subsubsection{Episodic Encoding}

ReViM retains contextual accounts of observed episodes and their established relations for reuse across questions. For a request \(a_t\) specifying a temporal interval, observation allocation and focus, the agent obtains \(o_t=\operatorname{Observe}(V,a_t)\) and updates memory as \(G_V^{t+1}=\operatorname{Write}(G_V^t,o_t)\).

Observe interprets frames and time-aligned subtitles or automatic speech recognition (ASR) text. Write retains the episode's interval, narrative, relations and original media support. Video descriptions persist across questions; tentative answers stay in the current reasoning state.

New episodes are immediately searchable. The observation record retains actual samples, modalities and sampling density, distinguishing broad coverage from detailed inspection. Later actions reuse these records or revisit their sources, keeping the active context bounded while adding local detail to persistent knowledge.

\subsubsection{Schema Association}

The memory graph \(G_V=(C,S,B)\) connects content records \(C\) through relation schemas \(S\). Content records include episodes and supported semantic records; schemas define relations and participant roles. Each binding in \(B\) links a record to a schema and records the participants, their role assignments and occurrence time. Episodes expressing the same relation can therefore share an access path without exchanging participants.

For example, two chase episodes share a schema but retain separate chaser and target bindings and their narrative context.

During writing, the agent aligns relation and role meanings with existing schemas, allowing equivalent descriptions to share an access path, or defines new schemas when meanings differ. The program registers each occurrence's participants, role assignments and time in its bindings. A reused schema requires a complete correspondence between observed and canonical roles.

As observations accumulate, the agent can integrate related episodes into semantic records, retaining their supporting episodes. Additional detail supplements an occurrence; a corrected interpretation revises it; a later state change remains a distinct occurrence. Revision withdraws dependent statements and experience cases. Both local inspection and broader temporal reasoning therefore operate on the same evolving memory.

\subsubsection{Relative Memory Retrieval}

A request \(x\) specifies an information need, grounded role or participant constraints and an optional interval. It admits the following content--schema associations:

\begin{equation}\label{eq:revim-3}
A_x(c,s)=\mathbf 1\!\left[
\exists b\in B(c,s):\operatorname{Compatible}(b,x)
\right].
\end{equation}

Here \(B(c,s)\) contains valid bindings between content \(c\in C\) and schema \(s\in S\). Compatibility jointly evaluates role and time conditions within one binding, preventing matches across different occurrences. Identity constraints require current grounding; uncertain identities remain descriptive cues. Returned episodes retain context and matching bindings.

Lexical matching and a frozen text encoder provide positive query restart weights \(r_x\) for content and schema nodes. Personalized PageRank~\cite{gutierrez2024hipporag} propagates activation on the allowed graph \(A_x\), extending access through shared schemas. Memory grows with earlier questions, so connectivity reflects past demand. A well-connected episode can receive high activation without being especially relevant; the neutral walk accounts for that structural contribution. A second walk therefore uses neutral weights \(r_0\), uniform within the same content and schema partitions. Both walks share transitions and propagation coefficient \(\alpha\), yielding \(p_z=\operatorname{PPR}(A_x,r_z;\alpha)\) for \(z\in\{x,0\}\). We score content by
\begin{equation}\label{eq:revim-4}
\mathrm{score}(c\mid x)=\log\frac{p_x(c)}{p_0(c)} .
\end{equation}

The ratio discounts activation explained by structural background. A direct-match quota preserves lexical and semantic hits; associated records fill the remaining return budget. Retrieved content and its supporting bindings enter \(E_t\), while subsequent observations extend \(G_V\).

\subsection{Retrieval Refinement}\label{sec:revim-retrieval-refinement}

\subsubsection{Reflecting on Retrieval}

A relevant return may still leave the required relation unresolved. ReViM responds to empty returns, repeated requests or consecutive retrievals without new content. The agent may also identify an unmet need during normal reasoning. Given the question, current evidence and request history \(H_t\), reflection produces

\begin{equation}\label{eq:revim-5}
(g_t,c_t,d_t,a_t^{\mathrm{loc}})
=\operatorname{Reflect}(q,E_t,H_t),
\end{equation}

where \(g_t\) specifies the missing information, \(c_t\) describes the current condition requiring adjustment, \(d_t\) states what the return must establish, and \(a_t^{\mathrm{loc}}\) is a local request. Goals and conditions use the same relations and roles as video memory. Known matches relations in \(E_t\) and binds their arguments; Unknown tests absence without supplying action inputs.

The requirement is fixed before the request executes. A request can revise its query focus or grounded constraints, expand a located episode, or inspect the video. In Fig.~\ref{fig:framework}, the known chase episode locates the rabbit, while its missing color calls for focused inspection of that occurrence.

\subsubsection{Conditional Experience Reuse}

Historical experience offers a complementary adjustment. A stored method \(m=(g,c,d,a)\) connects a generalized goal, applicability condition, completion check and action. The program matches its goal and check to the current declarations, then evaluates its condition on \(E_t\) to obtain a current argument map \(\theta_t\). For an applicable method \(m_t\), the resulting request is \(a_t^{\mathrm{exp}}=\operatorname{Instantiate}(m_t,\theta_t)\).

Arguments come from the current goal and established positive premises. The same map supplies current participants, relations, intervals and record addresses to both the condition and the action. For example, an experience for inspecting a located event is instantiated with the current event and its interval. The current completion requirement remains unchanged.

Applicable methods supported by resolved cases take priority. If only unresolved experience applies, its condition and previous action type guide the agent's local refinement. Without an applicable method, the controller executes its local request \(a_t^{\mathrm{loc}}\). Historical and local requests share the acquisition budget; unchanged-state repetitions are omitted. Each request retains its predeclared check. Returns update \(E_t\), and newly established video information updates \(G_V\). Executed refinements provide the cases learned next.

\subsection{Joint Experience Learning}\label{sec:revim-joint-experience-learning}

\subsubsection{Learning from Retrieval Outcomes}

To retain how missing information was acquired, ReViM extracts cases from actual adjustments. Each case preserves its goal, condition, completion requirement and executed action, with the before/after evidence and execution support. For an adjustment whose condition held and whose local target was initially unresolved, its outcome is

\begin{equation}\label{eq:revim-7}
\begin{aligned}
y_j&=\mathbf 1\!\left[d_j(E_j^{\mathrm{after}})=1\right],\\
&\quad d_j(E_j^{\mathrm{before}})=0.
\end{aligned}
\end{equation}

The predeclared check evaluates the actual return with the same before/after criterion. Resolved cases supply demonstrated actions; unresolved cases retain their condition and operation type for later reflection. This local feedback guides experience reuse, while the agent answers from accumulated video evidence. Both kinds retain their supporting records for subsequent revision.

\subsubsection{Joint Generalization}

Transferring an action requires preserving its connection to the need and conditions that justified it. We therefore generalize the four parts jointly. Let \(\tilde e_j\) expose the demonstrated action for a resolved case, or the condition and action type for an unresolved case:

\begin{equation}\label{eq:revim-8}
\begin{aligned}
\tilde e_j&=(g_j,c_j,d_j,a_j^{\mathrm{pub}}),\\
m&=\operatorname{Generalize}
       (\tilde e_1,\ldots,\tilde e_n).
\end{aligned}
\end{equation}

The program first groups cases with compatible operators and field structure, then recursively compares their four parts together. Every declared instance argument is lifted to a typed variable, including arguments whose values coincide across cases. Positions with the same value correspondence across all source cases share one variable; different correspondences remain separate. One variable table spans the goal, condition, check and action, preserving their execution dependencies. Fixed operators and public relation meanings retain their roles in applicability matching.

For a method that inspects a known participant's missing color, the participant variable links the goal and check to the inspection request; an episode established by the condition supplies its interval. At reuse, \(\theta_t\) binds these arguments to the current rabbit and chase episode (Fig.~\ref{fig:framework}). The color must be established by the new return.

\subsubsection{Experience Consolidation}

After prediction, eligible cases enter experience memory for subsequent videos. Compatible cases yield shared methods, prioritized by actual reuse outcomes and recency within a capacity budget. Revisions to supporting video knowledge withdraw affected cases and rebuild their methods. Subsequent executions continue the acquisition, assessment and reuse cycle.

\section{Experiments}
\label{sec:experiments}
\raggedbottom
\setlength{\intextsep}{8pt}

\setcounter{topnumber}{1}
\setcounter{dbltopnumber}{1}

\begin{table*}[!t]
\caption{\textbf{Comparison with end-to-end VLMs, video agents, and memory-based methods on four video understanding benchmarks.} All scores are accuracy (\%). $\dagger$ marks reproduced rows or entries; remaining results follow published reports. ReViM scores are bold, and green parentheses show percentage-point gains over the corresponding reproduced backbone.}
\label{tab:main-results}
\centering
\fontsize{8.5}{10}\selectfont
\setlength{\tabcolsep}{2.5pt}
\renewcommand{\arraystretch}{1.06}
\resizebox{\textwidth}{!}{%
\begin{tabular}{@{}l@{\hspace{3pt}}l@{\hspace{4pt}}r@{\hspace{2pt}}l@{\hspace{4pt}}r@{\hspace{2pt}}l@{\hspace{4pt}}r@{\hspace{2pt}}l@{\hspace{4pt}}r@{\hspace{2pt}}l@{\hspace{7pt}}r@{\hspace{2pt}}l@{\hspace{5pt}}}
\toprule
Method & Model(s) & \multicolumn{2}{c}{LVBench} & \multicolumn{4}{c}{LongVideoBench} & \multicolumn{2}{c}{Video-MME} & \multicolumn{2}{c}{EgoSchema} \\
\cmidrule(lr){5-8}
 & & \multicolumn{2}{c}{Overall} & \multicolumn{2}{c}{Val} & \multicolumn{2}{c}{Long} & \multicolumn{2}{c}{Long (w/o sub)} & \multicolumn{2}{c}{Val} \\
\midrule
\rowcolor{revimshade}
\multicolumn{12}{@{}l}{\textbf{End-to-End VLMs (Proprietary)}} \\
Gemini 2.5 Pro~\cite{comanici2025gemini} & Gemini 2.5 Pro & 72.0 &  & 71.0 &  & 68.6 &  & 75.9 &  & 72.8 &  \\
GPT-4o~\cite{hurst2024gpt} & GPT-4o & 48.9 &  & 66.7 &  & 60.9 &  & 65.3 &  & 70.4 &  \\
Seed1.5-VL~\cite{guo2025seed1} & Seed1.5-VL-Thinking & 64.6 &  & 74.0 &  & -- &  & -- &  & -- &  \\
GPT-5$^{\dagger}$ & GPT-5 & 60.4 &  & 65.7 &  & 61.9 &  & 74.3 &  & 74.4 &  \\
\midrule
\rowcolor{revimshade}
\multicolumn{12}{@{}l}{\textbf{End-to-End VLMs (Open-source)}} \\
Qwen2.5-VL-72B~\cite{bai2025qwen25vltechnicalreport} & Qwen2.5-VL-72B & 47.3 &  & 60.7 &  & 54.2 &  & 63.9 &  & 75.4 &  \\
InternVL2.5-78B~\cite{chen2024expanding} & InternVL2.5-78B & 43.6 &  & 63.6 &  & -- &  & 62.6 &  & -- &  \\
LLaVA-Video-72B~\cite{zhang2024llava} & LLaVA-Video-72B & 38.7 &  & 61.9 &  & -- &  & 61.5 &  & 74.7 &  \\
GLM-4.5V~\cite{hong2025glm} & GLM-4.5V-106B-A12B & 53.4 &  & 66.4 &  & 54.1 &  & 64.7 &  & 68.4 &  \\
AdaReTaKe~\cite{wang2025adaretake} & Qwen2.5-VL-72B & 53.3 &  & 67.0 &  & -- &  & 65.0 &  & -- &  \\
Qwen3-VL-8B$^{\dagger}$ & Qwen3-VL-8B & 48.5 &  & 62.6 &  & 52.1 &  & 61.2 &  & 72.0 &  \\
Qwen3.6-27B$^{\dagger}$ & Qwen3.6-27B & 65.5 &  & 70.9 &  & 61.5 &  & 65.6 &  & 73.8 &  \\
\midrule
\rowcolor{revimshade}
\multicolumn{12}{@{}l}{\textbf{Agentic and Memory-Based Methods}} \\
VideoAgent~\cite{wang2024videoagent} & GPT-4 & 29.3 &  & -- &  & 39.5 &  & 46.4 &  & 60.2 &  \\
VideoTree~\cite{wang2025videotree} & GPT-4 & 28.8 &  & -- &  & -- &  & 54.2 &  & 67.0 &  \\
Video-RAG~\cite{luo2026video} & LLaVA-Video-72B & -- &  & 65.4 &  & -- &  & 73.1 &  & -- &  \\
MR. Video~\cite{pang2025mrvideo} & GPT-4o & 60.8 &  & -- &  & 61.6 &  & 61.8 &  & 73.0 &  \\
DVD~\cite{zhang2026deep} & o3 & 74.2 &  & 71.6 &  & 68.6 &  & 67.3 &  & 76.6 &  \\
LensWalk~\cite{li2026lenswalk} & o3 + GPT-4.1 & 66.8 &  & -- &  & 69.9 &  & 70.0 &  & 74.8 &  \\
VideoLucy~\cite{zuo2026videolucy} & DeepSeek-R1 & 58.8 &  & 70.2 &  & -- &  & 66.8 &  & -- &  \\
WorldMM-8B~\cite{yeo2026worldmm} & Qwen3-VL-8B & 55.4$^{\dagger}$ &  & 62.4$^{\dagger}$ &  & 57.1$^{\dagger}$ &  & 66.0 &  & 71.4$^{\dagger}$ &  \\
MERIT-8B$^{\dagger}$~\cite{choi2026keep} & Qwen3-VL-8B & 56.5 &  & 63.8 &  & 58.2 &  & 66.2 &  & 73.0 &  \\
MM-Mem~\cite{lian2026verbatim} & Qwen3-VL-8B & -- &  & -- &  & -- &  & 66.1 &  & -- &  \\
M$^3$-Agent~\cite{long2026seeing} & Qwen3-32B & 49.3 &  & -- &  & -- &  & 61.8 &  & -- &  \\
AVP~\cite{wang2026active}$^{\dagger}$ & Qwen3.6-27B & 68.3 &  & 72.1 &  & 64.5 &  & 68.4 &  & 74.4 &  \\
VideoARM~\cite{yin2026videoarm}$^{\dagger}$ & Qwen3.6-27B & 67.5 &  & 72.3 &  & 63.8 &  & 69.2 &  & 73.6 &  \\
MemDreamer~\cite{chen2026memdreamer}$^{\dagger}$ & Qwen3.6-27B & 68.8 &  & 74.5 &  & 65.1 &  & 69.6 &  & 74.8 &  \\
WorldMM~\cite{yeo2026worldmm} & GPT-5 & 61.9 &  & -- &  & -- &  & 76.6 &  & -- &  \\
VideoSeek~\cite{lin2026videoseek} & GPT-5 & 68.4 &  & -- &  & 73.5 &  & 70.1 &  & -- &  \\
MERIT~\cite{choi2026keep} & GPT-5 & 71.8 &  & -- &  & -- &  & 77.7 &  & -- &  \\
\midrule
\revimname{} & Qwen3-VL-8B & \textbf{57.7} & \textcolor{revimgreen}{\scriptsize (+9.2)} & \textbf{65.9} & \textcolor{revimgreen}{\scriptsize (+3.3)} & \textbf{59.4} & \textcolor{revimgreen}{\scriptsize (+7.3)} & \textbf{67.8} & \textcolor{revimgreen}{\scriptsize (+6.6)} & \textbf{74.6} & \textcolor{revimgreen}{\scriptsize (+2.6)} \\
\revimname{} & Qwen3.6-27B & \textbf{69.1} & \textcolor{revimgreen}{\scriptsize (+3.6)} & \textbf{74.6} & \textcolor{revimgreen}{\scriptsize (+3.7)} & \textbf{65.4} & \textcolor{revimgreen}{\scriptsize (+3.9)} & \textbf{69.8} & \textcolor{revimgreen}{\scriptsize (+4.2)} & \textbf{75.2} & \textcolor{revimgreen}{\scriptsize (+1.4)} \\
\revimname{} & GPT-5 & \textbf{74.7} & \textcolor{revimgreen}{\scriptsize (+14.3)} & \textbf{75.8} & \textcolor{revimgreen}{\scriptsize (+10.1)} & \textbf{73.9} & \textcolor{revimgreen}{\scriptsize (+12.0)} & \textbf{78.4} & \textcolor{revimgreen}{\scriptsize (+4.1)} & \textbf{79.2} & \textcolor{revimgreen}{\scriptsize (+4.8)} \\
\bottomrule
\end{tabular}%
}
\end{table*}

\subsection{Benchmarks}
\label{sec:benchmarks}
We report accuracy on four multiple-choice benchmarks.
LVBench~\cite{wang2025lvbench} contains 1,549 questions over 103 hour-scale videos across six capability dimensions.
LongVideoBench~\cite{wu2024longvideobench} includes 1,337 validation questions (LVB-Val); its Long subset (LVB-Long) has 564 questions over 188 videos lasting 15--60 minutes.
Video-MME Long (VM-L)~\cite{fu2025video} comprises 900 questions over 300 videos of 30--60 minutes and is evaluated without provided subtitles.
EgoSchema~\cite{mangalam2023egoschema} tests reasoning over three-minute egocentric clips; we use its public 500-question Val subset.

\subsection{Experimental Setup}
\label{sec:experimental-setup}
\paragraph{Implementation.}
ReViM uses GPT-5~\cite{openai2025gpt5systemcard}, Qwen3.6-27B~\cite{qwen2026qwen36modelcard}, or Qwen3-VL-8B~\cite{bai2025qwen3} for perception and reasoning, with frozen BGE-M3 embeddings~\cite{chen-etal-2024-m3} and lexical matching for retrieval.
Broad intervals use sparse frames; local inspection uses denser sampling.
LVBench and LVB-Val obtain speech evidence from on-demand, time-aligned Whisper transcripts~\cite{pmlr-v202-radford23a}; LVB-Val excludes provided subtitles.
We use a maximum of $T_{\max}=5$ acquisition steps per question and set $\alpha=0.5$. Each Search, Read or Observe counts as one step, with early answering allowed.
Video memory persists across a video's questions and resets between videos.
Experience starts empty per benchmark. Eligible cases are committed after each prediction, and only preceding cases from other videos can be generalized and reused. Local acquisition outcomes provide feedback without reference answers.
Appendices~\ref{supp:implementation} and~\ref{supp:prompts} detail the evaluation settings and prompts.

\paragraph{Comparisons.}
Table~\ref{tab:main-results} combines results from the literature with our reproductions, marked with $\dagger$ at the row or entry level.
Controlled analyses use Qwen3-VL-8B. Variants follow the same fixed video and question order with independent memory states and retain the remaining model and budget settings.

\subsection{Main Results}
\label{sec:main-results}
Table~\ref{tab:main-results} shows improvements over all three matched backbones.
GPT-5 gains 14.3 points on LVBench, 10.1/12.0 on LVB-Val/LVB-Long, 4.1 on VM-L, and 4.8 on EgoSchema Val, yielding the highest scores among the listed configurations.
Qwen3-VL-8B and Qwen3.6-27B gain 2.6--9.2 and 1.4--4.2 points, respectively.
The 27B configuration exceeds our AVP, VideoARM and MemDreamer reproductions across all five metrics; the 8B configuration exceeds WorldMM-8B and MERIT-8B.
The gains across backbones and video domains support persistent knowledge and focused acquisition; the component analyses below examine their contributions.

\paragraph{End-to-end VLMs.}
With GPT-5, ReViM exceeds Gemini 2.5 Pro by 2.7 points on LVBench and 5.3 on LVB-Long. Its 8B configuration also exceeds Qwen2.5-VL-72B by 10.4 points on LVBench. These comparisons show gains beyond the matched-backbone setting.

\paragraph{Memory-based retrieval.}
WorldMM~\cite{yeo2026worldmm} retrieves multimodal memories across temporal scales, while MERIT~\cite{choi2026keep} expands multi-key episodic retrieval with neighboring clips. With the same 8B backbone, ReViM gains 2.3/1.2 points on LVBench and 3.5/2.1 on LVB-Val over these methods. ReViM organizes observations acquired during questions through shared schemas and occurrence bindings, then retains subsequent refinements, connecting current retrieval with continued knowledge formation.

\paragraph{Online agentic acquisition.}
AVP~\cite{wang2026active} iteratively plans, observes and reflects, while VideoARM~\cite{yin2026videoarm} couples adaptive tools with hierarchical memory. Our 27B configuration improves over their reproduced results by 0.8/1.6 points on LVBench and 2.5/2.3 on LVB-Val. ReViM connects local acquisition adjustments with persistent knowledge and transferable procedures.

\begin{figure*}[!t]
\centering
\includegraphics[width=0.653\textwidth]{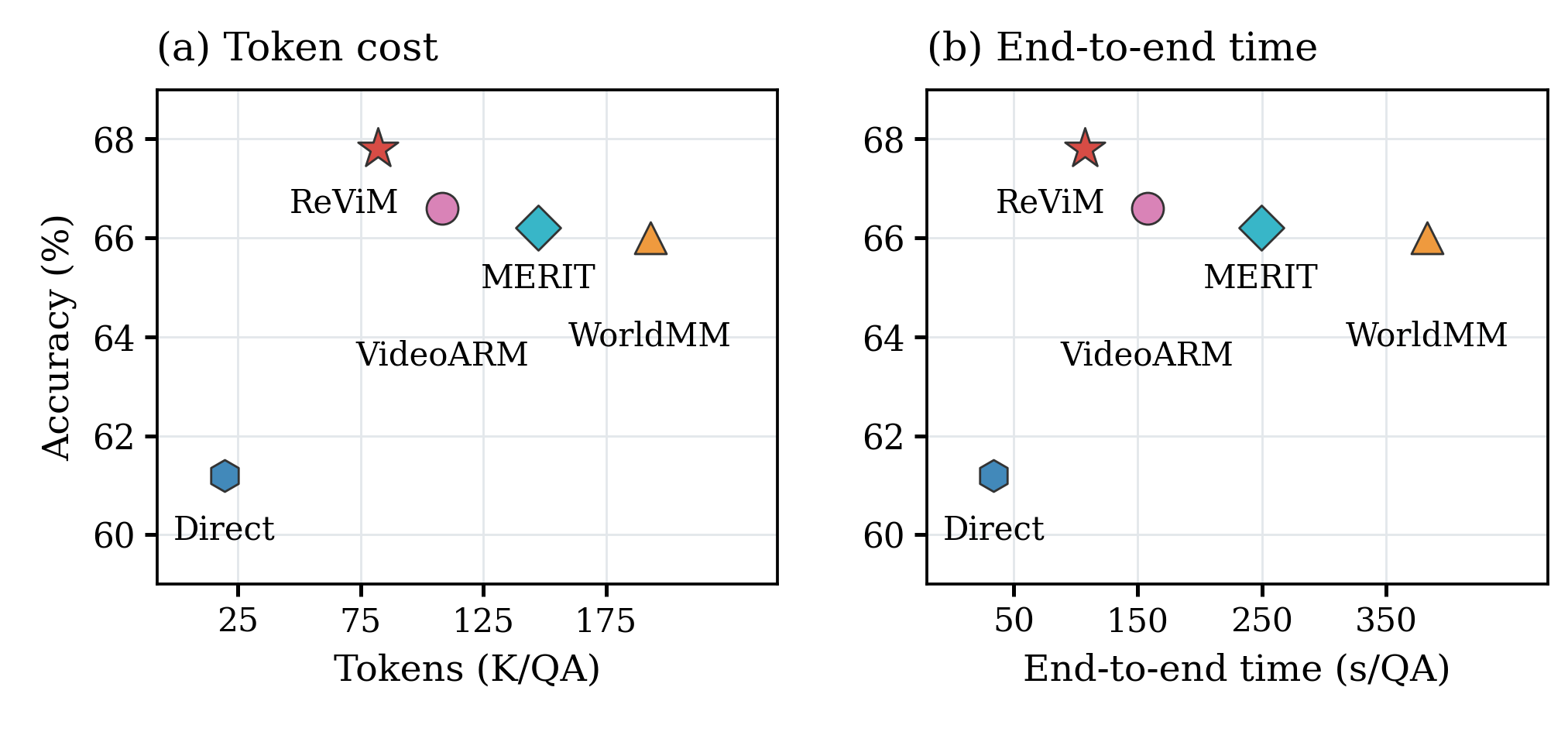}\hfill
\includegraphics[width=0.327\textwidth]{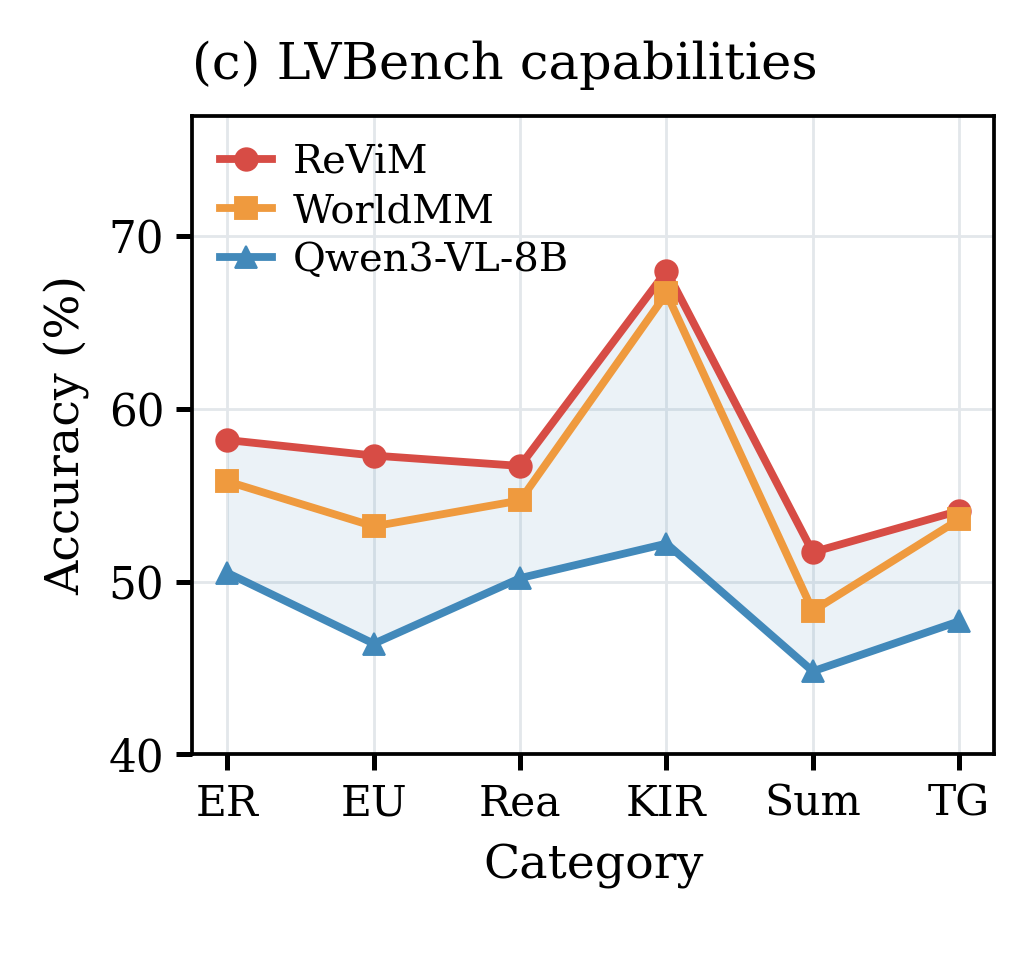}
\caption{\textbf{Efficiency and capability analysis with Qwen3-VL-8B.} (a,b) VM-L accuracy versus cumulative tokens and end-to-end time per QA, including preprocessing and memory updates. (c) LVBench accuracy across six capability categories; shading shows the gap between ReViM and the backbone. ER: entity recognition; EU: event understanding; Rea: reasoning; KIR: key information retrieval; Sum: summarization; TG: temporal grounding.}
\label{fig:resource-efficiency}
\label{fig:lvbench-types}
\end{figure*}

Figure~\ref{fig:lvbench-types}(c) further breaks down LVBench performance by capability. ReViM improves over the reproduced backbone in all six categories, with the largest gain in key information retrieval (+15.8 points). Improvements in event understanding (+10.9) and reasoning (+6.5) also indicate benefits beyond recognizing isolated entities, consistent with retaining episode context and acquiring missing relations.
Compared with WorldMM, gains in event understanding and summarization reach 4.1 and 3.4 points, whereas temporal grounding improves by 0.5 points. This pattern suggests that ReViM's advantage extends from locating relevant moments to retaining and connecting their semantic content.

\subsection{Ablation Studies}
\label{sec:component-analysis}

Table~\ref{tab:ablations} tests acquisition refinement and memory organization.
\emph{Memory only} retains associative memory and ordinary agent reasoning and tools, disabling structured refinement and cross-video experience.
\emph{w/o cross-video experience} retains video memory and local refinement, disabling historical guidance.
Local refinement adds 1.8/1.5/2.5 points; cross-video experience adds 2.2/1.1/1.6.
\emph{Reset video memory per question} clears video memory between questions, retaining cross-video experience; its 2.3--3.2-point loss supports cross-question accumulation.

\begin{table}[!ht]
\caption{\textbf{Ablation of memory organization and acquisition refinement.} Accuracy (\%) with Qwen3-VL-8B. Each variant removes or replaces the indicated component; shading marks Full ReViM.}
\label{tab:ablations}
\centering
\small
\setlength{\tabcolsep}{3pt}
\begin{tabular}{@{}lrrr@{}}
\toprule
Setting & VM-L & LVB-Val & LVBench \\
\midrule
Memory only & 63.8 & 63.3 & 53.6 \\
w/o cross-video experience & 65.6 & 64.8 & 56.1 \\
Reset video memory per question & 64.6 & 63.6 & 55.0 \\
w/o shared schemas & 65.4 & 64.5 & 55.7 \\
w/o binding constraints & 65.9 & 65.1 & 56.4 \\
Absolute activation & 66.3 & 65.2 & 56.8 \\
\rowcolor{revimourshade}
Full ReViM & \textbf{67.8} & \textbf{65.9} & \textbf{57.7} \\
\bottomrule
\end{tabular}
\end{table}

\emph{w/o shared schemas} uses episode-local relation entries with unchanged labels, narrative records and role bindings.
\emph{w/o binding constraints} retains the graph and bindings for encoding and experience instantiation but bypasses retrieval compatibility filtering (Eq.~\ref{eq:revim-3}).
Other mechanisms remain unchanged.
Their respective 1.4--2.4 and 0.8--1.9-point losses support cross-episode association and occurrence-specific retrieval.

\emph{Absolute activation} replaces the score in Eq.~\ref{eq:revim-4} with $p_x(c)$, preserving construction, query seeding, the propagation coefficient and return budgets.
Its 0.7--1.5-point loss supports ranking relative to structural background.
The schema and binding ablations link the gains to cross-episode association and occurrence-specific access. Together with the reset variant, they support retaining both episode context and its organization across questions.

\subsection{Experience Transfer}
\label{sec:experience-transfer}

Joint Experience Learning (JEL) uses adapters to expose each host's information needs, executed requests and local returns, retaining its factual memory, tools and acquisition budget.
Table~\ref{tab:experience-transfer} shows gains of 1.1/1.2 points on LVB-Val and 0.9/1.0 on VM-L.
Improvements across both hosts support portability beyond ReViM's memory graph. Under the same acquisition budgets, the transferred procedures improve the use of each host's existing memory and tools.

\begin{table}[H]
\caption{\textbf{Transfer of Joint Experience Learning (JEL) to existing video memory frameworks.} Accuracy (\%) with Qwen3-VL-8B. Green parentheses give gains over each original host.}
\label{tab:experience-transfer}
\centering
\small
\setlength{\tabcolsep}{3pt}
\begin{tabular}{@{}lr@{\,}lr@{\,}l@{}}
\toprule
Method & \multicolumn{2}{c}{LVB-Val} & \multicolumn{2}{c}{VM-L} \\
\midrule
WorldMM & 62.4 & & 66.0 & \\
WorldMM + JEL & \textbf{63.5} & \textcolor{revimgreen}{\scriptsize (+1.1)} & \textbf{66.9} & \textcolor{revimgreen}{\scriptsize (+0.9)} \\
MERIT & 63.8 & & 66.2 & \\
MERIT + JEL & \textbf{65.0} & \textcolor{revimgreen}{\scriptsize (+1.2)} & \textbf{67.2} & \textcolor{revimgreen}{\scriptsize (+1.0)} \\
\bottomrule
\end{tabular}
\end{table}

\subsection{Parameter Sensitivity}
\label{sec:parameter-sensitivity}

Table~\ref{tab:parameter-sensitivity} varies the association weight and maximum acquisition steps one at a time, with the remaining settings fixed.
$\alpha=0.5$ improves over direct seeding ($\alpha=0$) by 0.9--2.1 points.
Increasing $T_{\max}$ from 1 to 5 gains 3.4--4.3 points; extending to 7 changes accuracy by $-1.2$ to $+0.3$ points.
The gain from one to three steps exceeds that from three to five on every benchmark, showing diminishing returns as the acquisition budget grows.
We therefore set $T_{\max}=5$ across benchmarks.

\begin{table}[!ht]
\caption{\textbf{Sensitivity to the propagation coefficient and acquisition budget.} Accuracy (\%) with Qwen3-VL-8B. We vary $\alpha$ and the maximum acquisition steps $T_{\max}$ separately; shading marks defaults.}
\label{tab:parameter-sensitivity}
\centering
\small
\setlength{\tabcolsep}{6pt}
\begin{tabular}{@{}lrrr@{}}
\toprule
Setting & VM-L & LVB-Val & LVBench \\
\midrule
$\alpha=0$ & 65.7 & 65.0 & 56.6 \\
$\alpha=0.25$ & 66.3 & 65.4 & 57.5 \\
\rowcolor{revimourshade}
$\alpha=0.5$ & 67.8 & 65.9 & 57.7 \\
$\alpha=0.75$ & 66.2 & 65.8 & 57.4 \\
\midrule
$T_{\max}=1$ & 63.6 & 62.5 & 53.4 \\
$T_{\max}=3$ & 66.0 & 65.1 & 56.5 \\
\rowcolor{revimourshade}
$T_{\max}=5$ & 67.8 & 65.9 & 57.7 \\
$T_{\max}=7$ & 66.6 & 66.2 & 57.6 \\
\bottomrule
\end{tabular}
\end{table}

\subsection{Efficiency}
\label{sec:efficiency}

Figure~\ref{fig:resource-efficiency}(a,b) compares ReViM with WorldMM, MERIT and VideoARM, and includes the direct backbone reference. The backbone obtains 61.2\% at 19.6K tokens and 34.0 seconds per QA.
ReViM reaches 67.8\% with 82.2K tokens and 107.1 seconds per QA.
Against WorldMM, MERIT and VideoARM, it reduces tokens by 57.4\%, 44.3\% and 24.2\%, with respective end-to-end speedups of 3.58$\times$, 2.33$\times$ and 1.47$\times$.

\begin{table}[!t]
\caption{\textbf{Average resource use per question for ReViM.} Qwen3-VL-8B; tokens (K) are grouped by processing stage. Frames and Calls count frame presentations and model generation calls, respectively.}
\label{tab:internal-cost}
\centering
\small
\setlength{\tabcolsep}{2.5pt}
\begin{tabular}{@{}lrr@{}}
\toprule
Resource & VM-L & LVB-Val \\
\midrule
Frames & 162 & 96 \\
Calls & 8 & 7 \\
\midrule
Visual observation & 38.5 & 29.3 \\
Episodic and semantic encoding & 22.7 & 17.1 \\
Reasoning and answering & 21.0 & 14.9 \\
\rowcolor{revimourshade}
Total tokens & \textbf{82.2} & \textbf{61.3} \\
\bottomrule
\end{tabular}
\end{table}

Table~\ref{tab:internal-cost} shows that visual observation accounts for approximately 47\% of total tokens on both datasets, followed by memory encoding. Frames count presentations, including repeated inspection. Generation calls cover perception, encoding, retrieval refinement and answering. Tokens sum input and output across calls, including experience context; each call is counted once. End-to-end time includes graph retrieval and experience generalization.

ReViM exceeds WorldMM, MERIT and VideoARM by 1.8, 1.6 and 1.2 points, respectively, while reducing both tokens and end-to-end time. Together with the ablation and transfer results, these comparisons support persistent knowledge and reusable acquisition procedures as complementary ways to improve the use of limited observation budgets. Host transfer further supports reuse beyond ReViM's memory graph.

\section{Conclusion}
\label{sec:conclusion}
We introduced ReViM, a training-free memory lifecycle connecting on-demand video knowledge, retrieval refinement and cross-video experience. Associative memory links episodes through shared schemas and occurrence bindings. Refinement acquires missing information, while joint generalization transfers acquisition procedures. Experiments across four benchmarks and three backbones, with ablation, transfer and efficiency analyses, support persistent video knowledge and reusable experience. Both memories guide future retrieval. Experience guides acquisition in subsequent videos. \label{page:main-end}

\FloatBarrier

{
  \small
  \bibliographystyle{ieeenat_fullname}
  \bibliography{main,supplement_refs}
}
\input{sec/X_suppl.tex}
\end{document}

%% file: sec/X_suppl.tex
\clearpage
\appendix
\twocolumn[{\centering\Large\bfseries Appendix\par\vspace{0.8em}}]
\setcounter{table}{0}
\setcounter{figure}{0}
\renewcommand{\thefigure}{S\arabic{figure}}
\setcounter{equation}{0}
\renewcommand{\thetable}{S\arabic{table}}
\renewcommand{\theequation}{S\arabic{equation}}
\renewcommand{\theHfigure}{appendix.\arabic{figure}}
\renewcommand{\theHtable}{appendix.\arabic{table}}
\renewcommand{\theHequation}{appendix.\arabic{equation}}

\newcounter{revimalgorithm}
\renewcommand{\therevimalgorithm}{S\arabic{revimalgorithm}}
\newenvironment{revimalgorithm}[1]{%
  \begin{figure}[!htbp]\refstepcounter{revimalgorithm}
  \hrule\vspace{3pt}\noindent\textbf{Algorithm \therevimalgorithm. #1}\par
  \vspace{3pt}\hrule\vspace{4pt}\small\raggedright\setlength{\parindent}{0pt}
}{\par\vspace{3pt}\hrule\end{figure}}
\definecolor{suppblue}{RGB}{40,89,145}
\definecolor{suppteal}{RGB}{24,119,111}
\definecolor{supporange}{RGB}{151,91,31}
\definecolor{supppurple}{RGB}{112,76,151}
\definecolor{suppgray}{RGB}{90,100,112}
\definecolor{suppresultshade}{RGB}{232,244,238}
\newcommand{\algstage}[2][0]{\par\smallskip\algline{#1}{\textcolor{supppurple}{\bfseries #2}}\nobreak}
\newcommand{\algline}[2]{\par\begingroup\advance\leftskip by #1em\relax
  \hangindent=.8em\hangafter=1\noindent #2\par\endgroup}
\newcommand{\reviminput}[1]{\textcolor{suppblue}{\{#1\}}}
\newcommand{\promptpart}[1]{\par\smallskip\noindent\textbf{#1}\par}
\lstdefinestyle{revimresponse}{basicstyle=\ttfamily\fontsize{9.2}{11.4}\selectfont,
  columns=fullflexible,keepspaces=true,breaklines=true,showstringspaces=false,
  morestring=[b]",stringstyle=\color{suppblue},aboveskip=3pt,belowskip=2pt}
\newcommand{\revimpromptfont}{\fontsize{9.8}{11.4}\selectfont}
\newenvironment{revimprompt}[3][suppblue]{%
  \par\smallskip\noindent\begin{minipage}{\linewidth}%

  \captionsetup{hypcap=false,type=figure}%
  \def\revimpromptcaption{\caption{#2.}\label{#3}}%
  \begin{tcolorbox}[enhanced,colback=#1!3!white,colframe=#1!65!black,
    colbacktitle=#1!15!white,coltitle=black,title={#2},fonttitle=\bfseries,
    arc=3pt,boxrule=0.5pt,left=5pt,right=5pt,top=4pt,bottom=4pt,
    before skip=0pt,after skip=0pt]
  \normalfont\revimpromptfont\raggedright
}{%
  \end{tcolorbox}\revimpromptcaption\end{minipage}\par\smallskip
}
\begingroup
\linespread{.96}\selectfont
\setlength{\parskip}{0pt plus 1pt}
\section{Additional Method Details}
\label{supp:method}
\subsection{Memory and Execution}
\label{supp:execution}
Video memory $G_V$ links each episode to its source interval and participant bindings. For each question, retrieved records and new observations contribute to the working evidence $E_t$. When an occurrence is corrected, the program withdraws statements and experience cases that depend on it. A later change of state creates a separate occurrence.

\begin{revimalgorithm}{Continuous memory lifecycle}
\label{supp:alg:lifecycle}
\algline{0}{\textbf{Input} Video $V$, ordered questions, cases $\mathcal D$, step budget $T_{\max}$.}
\algline{0}{\textbf{Output} Predictions, video memory $G_V$, updated cases $\mathcal D$.}
\algstage{Acquire evidence and refine requests}
\algline{0}{Initialize $G_V$ for video $V$.}
\algline{0}{\textbf{for each} question $q_k$ \textbf{do}}
\algline{1}{Clear the evidence $E$, history $H$, and pending cases $\mathcal C_q$. Select eligible cases from preceding videos.}
\algline{1}{\textbf{while} acquisition budget remains \textbf{do}}
\algline{2}{If the evidence suffices, answer and exit the loop. Otherwise queue the request from $q_k,E,H$.}
\algline{2}{For refinement, fix $(g,c,d,a^{\rm loc})$ and schedule local and historical requests (Alg.~\ref{supp:alg:jel}).}
\algline{2}{Execute pending requests within budget. Skip unchanged-state repeats and refinements whose need is already met. Update $E,H$ after each return and write new observations to $G_V$.}
\algline{2}{If $c(E^-)=1$ and $d(E^-)=0$ before an adjustment, store the case in $\mathcal C_q$ with its post-return outcome $d(E^+)$.}
\algline{1}{At the limit, answer from $E,H$. After prediction, commit $\mathcal C_q$ and retain $G_V$ for the next question.}
\end{revimalgorithm}

\subsection{Relative Memory Retrieval}
\label{supp:retrieval}
We keep content--schema edges with compatible occurrence bindings, add self-loops to isolated nodes, and normalize each transition row. Reciprocal-rank fusion (offset 60) combines lexical and dense ranks. Both restarts assign equal mass to each nonempty node type. Within each type, the query restart follows the normalized fusion weights and the neutral restart is uniform. Both walks use the same transitions and $\alpha$. Direct hits take reserved slots; relative activation fills the rest after deduplication (Table~\ref{supp:tab:configuration}).

\subsection{Joint Experience Learning}
\label{supp:jel}
\paragraph{Shared arguments.}
META~\cite{huang2026meta} composes recurring tool sequences into macro-tools. ReViM generalizes each adjustment together with its goal, condition and completion check.
The program evaluates the same check $d$ before and after an adjustment. When the return meets the previously unmet need, $a^{\rm pub}$ contains the executed request. Otherwise, that field contains only its operation type. One variable table spans $(g,c,d,a^{\rm pub})$ across compatible cases. Two argument positions share a variable when their types and source identities agree in every case. Structural operators and public role meanings remain fixed.

\begin{revimalgorithm}{Joint generalization and instantiation}
\label{supp:alg:jel}
\algline{0}{\textbf{Input} Eligible cases, current declaration $(g_t,c_t,d_t,a_t^{\rm loc})$, and evidence $E_t$.}
\algline{0}{\textbf{Output} A request with the current completion check.}
\algstage{Generalize across source cases}
\algline{0}{Group public projections $(g,c,d,a^{\rm pub})$ by compatible structure and local outcome.}
\algline{0}{\textbf{for each} group \textbf{do}}
\algline{1}{Traverse the four fields recursively. Lift each declared instance argument to a typed variable using the shared table.}
\algline{1}{Recover each case's four fields by substituting its source values.}
\algstage{Bind to the current video}
\algline{0}{\textbf{for each} candidate method $m$ \textbf{in priority order do}}
\algline{1}{Match its goal and check to $g_t,d_t$ to seed $\theta$. Extend $\theta$ with bindings from positive Known premises in $E_t$.}
\algline{1}{Evaluate Unknown tests under $\theta$. An absent value leaves the need open and supplies no action argument.}
\algline{1}{\textbf{if} the match or condition fails: \textbf{continue}.}
\algline{1}{\textbf{if} $m$ is resolved and all action arguments are bound \textbf{then}}
\algline{2}{Substitute current participants, source intervals, and record addresses into the request using $\theta$.}
\algline{2}{\textbf{return} $\operatorname{Instantiate}(m,\theta)$ with the current check $d_t$.}
\algline{1}{Retain the condition and action type of unresolved matches.}
\algline{0}{\textbf{if} only diagnostic feedback applies, use it to revise $a_t^{\rm loc}$. \textbf{return} the local request with $d_t$.}
\end{revimalgorithm}

\paragraph{Binding example.}
Binding $x$ to \texttt{rabbit\_2} and $I$ to $[120,128]$ seconds directs Observe to inspect that rabbit in $I$. The check asks whether the return establishes a color for $x$ in $I$. Encoding blue for \texttt{rabbit\_2} with frame \texttt{F003} as support satisfies that check.

\section{Implementation and Evaluation Details}
\label{supp:implementation}
\subsection{Model Roles and Input Processing}
\label{supp:model-roles}
A single backbone interprets observations, selects actions, encodes relations and answers questions. BGE-M3 embeds text for retrieval. The program propagates graph scores, evaluates conditions and generalizes experience. For the two Qwen backbones, we sample with temperature 0.3 and top-$p$ 0.8.

\paragraph{Visual and speech input.}
Surveys sample across the requested interval. Focused inspections concentrate frames within a selected interval. Resolution adapts to the frame and visual-token budgets in Table~\ref{supp:tab:configuration}. LVBench and LongVideoBench use time-aligned Whisper transcripts. LongVideoBench excludes provided subtitles.

\begin{table}[t]
\caption{\textbf{ReViM evaluation settings.} Caps apply per observation or per question as indicated.}
\label{supp:tab:configuration}
\renewcommand{\arraystretch}{1.14}\centering\small\setlength{\tabcolsep}{3pt}
\begin{tabular}{@{}lr@{}}
\toprule
Setting&Value \\
\midrule
Acquisition steps per question ($T_{\max}$)&5 \\
Propagation coefficient ($\alpha$)&0.5 \\
Records per Search / reserved direct hits&6 / up to 3 \\
\midrule
Source frames per survey / local inspection&128 / 32 \\
Source-frame presentations / visual tokens per QA&320 / 131,072 \\
Text budget for retrieved records (tokens)&6,000 \\
\midrule
Active source cases in experience memory&256 \\
\bottomrule
\end{tabular}
\end{table}

\subsection{Budgets and Sequential Evaluation}
\label{supp:protocol}
\paragraph{Execution and memory.}
Search, Read and Observe each consume one acquisition step. The agent may answer early. Every 16 questions or on overflow, maintenance selects cases by support, reuse outcomes and recency, with a grace period for new cases.

\paragraph{Evaluation order.}
We process videos in order of first appearance and preserve annotation order within each video. Video memory resets between videos. Experience memory starts empty for each benchmark. Only cases committed after prediction on earlier videos are eligible for reuse. Reference answers are used for scoring. The LongVideoBench Long score uses 564 predictions from the Val stream. Across the default order and two shuffles of the videos, Qwen3-VL-8B gains 1.2--1.6 points on LVBench and 0.9--1.1 on LVB-Val over disabling experience in the corresponding order.

\paragraph{Comparisons and cost.}
On VM-L, Direct and ReViM use the same checkpoint and may transcribe source audio. Both exclude provided subtitles. Direct answers independently. ReViM retains video memory and can reopen source intervals. In the transfer experiment, WorldMM and MERIT expose their retrieval requests and returns to JEL and keep their own tools and budgets.

Efficiency comparisons use the 8B backbone. Token counts sum inputs and outputs across calls, including experience context. Video-level costs are amortized across questions, and repeated frames count on each presentation. Time includes decoding, ASR, generation, retrieval and memory updates. On VM-L, ReViM gains 6.6 accuracy points over Direct at $4.19\times$ the tokens and $3.15\times$ the time per question. It achieves higher accuracy with fewer tokens and less time than WorldMM, MERIT and VideoARM.

\FloatBarrier
\section{Additional Experiments}
\label{supp:experiments}
We use Qwen3-VL-8B throughout this section. VM-L denotes Video-MME Long without provided subtitles, and LVB-Val denotes the full LongVideoBench validation set.

\subsection{Alternative Experience Reuse Strategies}
\label{supp:experience-variants}
With retrieved cases, the agent adapts an individual example. With textual workflows, it follows a procedure summarized from prior cases. All variants use the same acquisition step limit.

\begin{table}[!htbp]
\caption{\textbf{Experience reuse strategies.} Accuracy (\%) with Qwen3-VL-8B. All variants retain video memory and local refinement.}
\label{supp:tab:experience-variants}
\centering\small\setlength{\tabcolsep}{4pt}
\begin{tabular}{@{}lrr@{}}
\toprule
Setting&LVBench&LVB-Val \\
\midrule
w/o cross-video experience&56.1&64.8 \\
Retrieved case reuse&56.6&65.0 \\
Textual workflow reuse&56.4&65.2 \\
\rowcolor{suppresultshade}
Full ReViM&\textbf{57.7}&\textbf{65.9} \\
\bottomrule
\end{tabular}
\end{table}

Full ReViM exceeds the strongest alternative by 1.1 percentage points on LVBench and 0.7 on LVB-Val (Table~\ref{supp:tab:experience-variants}).

\paragraph{Generalization strategy.}
Field-wise generalization matches arguments by type and role after generalizing each field independently. Standard joint generalization processes the fields together and keeps a value constant if it agrees at that position across cases. Full JEL lifts even these values to typed variables when they occupy declared instance slots. Positions share a variable when their types and identities agree within every source case (Alg.~\ref{supp:alg:jel}).

We sample 200 questions across videos in proportion to the six LVBench category frequencies. Earlier questions are processed in their original order to build memory, and only the sampled questions are scored. For binding, we sample 50 decisions from these questions, taking the first state per question with a declared unmet need and available historical cases. All strategies receive the same evidence and cases at each state before choosing a request.

\begin{table}[!htbp]
\caption{\textbf{Argument generalization.} QA accuracy on 200 LVBench questions. Binding counts use the same 50 reuse decisions across strategies.}
\label{supp:tab:argument-generalization}
\centering\small\setlength{\tabcolsep}{3pt}
\begin{tabular}{@{}lrrrr@{}}
\toprule
& & \multicolumn{3}{c}{Binding outcomes (counts)} \\
\cmidrule(l){3-5}
Strategy & QA (\%) & Correct & Incorrect & No request \\
\midrule
Field-wise & 56.5 & 34 & 9 & 7 \\
Standard joint & 57.5 & 38 & 6 & 6 \\
\rowcolor{suppresultshade}
Full JEL & \textbf{58.5} & \textbf{44} & 3 & 3 \\
\bottomrule
\end{tabular}
\end{table}

A correctly bound request matches every required argument to the intended entity, event or interval in the current evidence. Any mismatch in a complete request is incorrect. \emph{No request} indicates that no complete request was produced. In 50 reuse decisions, Full JEL produces 44 correctly bound requests, compared with 38 for standard joint and 34 for field-wise generalization. It improves QA accuracy over standard joint by 1.0 percentage point (Table~\ref{supp:tab:argument-generalization}).

\subsection{Video Memory and Cross-Video Experience}
\label{supp:factorial-memory}
We vary whether video memory persists across questions and whether the agent can reuse experience from earlier videos. In the reset setting, each question starts with empty video memory.

\begin{table}[!htbp]
\caption{\textbf{Video-memory persistence and cross-video experience.} Accuracy (\%); question-local refinement is retained in every setting.}
\label{supp:tab:factorial-memory}
\centering\small\setlength{\tabcolsep}{3pt}
\begin{tabular}{@{}llrrr@{}}
\toprule
Video memory&Experience&VM-L&LVB-Val&LVBench \\
\midrule
\rowcolor{suppresultshade}
Shared&On&\textbf{67.8}&\textbf{65.9}&\textbf{57.7} \\
Shared&Off&65.6&64.8&56.1 \\
Reset&On&64.6&63.6&55.0 \\
Reset&Off&62.2&61.9&53.3 \\
\bottomrule
\end{tabular}
\end{table}

Across the three benchmarks, retaining video memory adds 2.8--3.4 points with experience disabled. When video memory resets at each question, cross-video experience adds 1.7--2.4 points (Table~\ref{supp:tab:factorial-memory}).

\subsection{Cost as Questions Accumulate}
\label{supp:cross-question-cost}
We fix a cohort of LVBench videos with at least twelve questions and follow annotation order within each video. At prefix $m$, we divide the generation tokens spent on the first $m$ questions of each video by $m$, then average over videos.

\begin{figure}[!htbp]
\centering
\includegraphics[width=\linewidth]{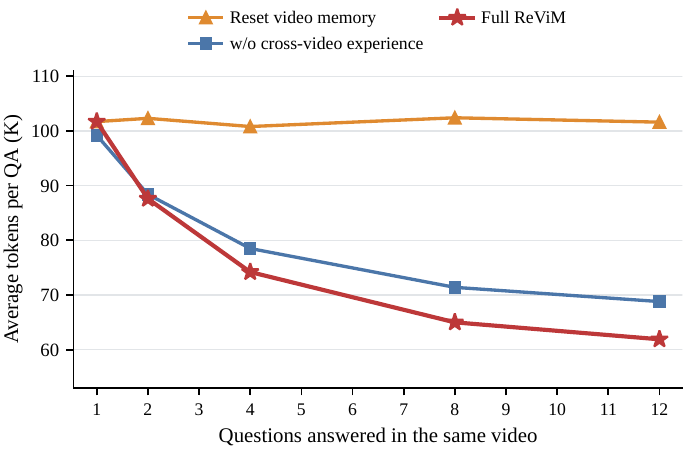}
\caption{\textbf{Cost of successive questions on the same video.} Cumulative average generation tokens per QA on a fixed LVBench video cohort. Markers indicate evaluated prefixes.}
\label{supp:fig:cross-question-cost}
\end{figure}

Full ReViM and the reset variant both start at 101.7K tokens per QA. Across twelve questions, their averages are 61.9K and 101.6K. Keeping video memory with experience disabled gives 68.8K (Fig.~\ref{supp:fig:cross-question-cost}).

\subsection{Temporal Evidence Retrieval}
\label{supp:temporal-retrieval}
We use 200 LVBench questions whose annotated event is already represented in video memory. All variants receive the same question-start snapshots, queries and grounded constraints, with the same six-record limit and text budget. Hybrid retrieval combines lexical and dense ranks directly. We score each question by the best temporal intersection over union (tIoU) with the reference, assigning zero to an empty return. Recall@6 is the fraction reaching each threshold.

\begin{table}[!htbp]
\caption{\textbf{Temporal evidence retrieval from fixed memory states.} Mean best tIoU and Recall@6 are percentages; Span is the average returned source-interval union in seconds.}
\label{supp:tab:temporal-retrieval}
\renewcommand{\arraystretch}{1.12}\centering\small\setlength{\tabcolsep}{3pt}
\begin{tabular}{@{}lrrrrr@{}}
\toprule
&Best&\multicolumn{3}{c}{Recall@6}&Span \\
\cmidrule(lr){3-5}
Variant&tIoU&0.1&0.3&0.5&(s) \\
\midrule
Hybrid retrieval&29.1&68.0&43.5&21.0&252 \\
Absolute activation&31.4&72.0&47.5&25.0&246 \\
w/o binding constraints&32.2&74.0&50.0&26.0&249 \\
\rowcolor{suppresultshade}
ReViM retrieval&\textbf{35.6}&\textbf{79.0}&\textbf{55.5}&\textbf{31.5}&242 \\
\bottomrule
\end{tabular}
\end{table}

At tIoU 0.3, ReViM retrieves a matching interval for 111 of 200 questions, compared with 95 under absolute activation. The returned intervals cover 242 and 246 seconds per question on average, respectively (Table~\ref{supp:tab:temporal-retrieval}).

\FloatBarrier
\section{Limitations and Discussion}
\label{supp:limitations}
Sparse samples can identify an event while leaving its order uncertain. In Fig.~\ref{supp:fig:first-instrument}, the survey shows the stage and then a bass, missing the intervening drum kit. The episode calls the bass the first instrument, and the controller accepts this account. Reopening the frame verifies the sighting. Checking the order requires inspecting the preceding gap. The scope of the claim could guide that decision.

ReViM uses a single backbone to interpret footage, encode relations and decide when the evidence is sufficient to answer. These roles place different demands on the model. In the diving run without experience (Fig.~\ref{supp:fig:board-diving}), the reader identifies Germany's third dive and China's lead after round three. The controller nevertheless uses commentary about the German pair to answer a question about the champions. This distinction suggests a way to study backbone choice: hold the retrieved footage fixed and vary the models used to describe it and to reason over the resulting account.

Revisiting episodes requires access to their footage. Q3 reopens Q1's scoreboard to verify the scoring rule (Fig.~\ref{supp:fig:board-diving}). In a live stream, the episode may outlast the clip. Acquisition procedures can remain useful across videos, as the handshake case illustrates in the diving question (Fig.~\ref{supp:fig:experience-return}). Future work could examine how to select source clips for later questions and retain acquisition procedures that remain useful as video content changes.

\endgroup
\section{Prompt Templates}
\label{supp:prompts}
Figures~\ref{supp:fig:controller}--\ref{supp:fig:perception} give condensed prompts for ReViM and the Direct baseline.

\begin{revimprompt}[suppblue]{Controller and retrieval refinement}{supp:fig:controller}
\promptpart{Available state}
\reviminput{question and options}, \reviminput{read records and sources}, \reviminput{request history, sampling and budget}, \reviminput{current need and feedback}.

\promptpart{Refine a request}
Answer when the evidence distinguishes the options. Otherwise specify the missing relation, participant and interval. Anchor a located event to its read record. If a request repeats an earlier return, change the query, read the episode or inspect its source. Bind historical guidance to current inputs. At the step limit, answer from the evidence collected.
\promptpart{Example action field}
\begin{lstlisting}[style=revimresponse]
{"kind":"Read",
 "record_ids":["c000003@1"],
 "material":"media",
 "focus":"inspect target"}
\end{lstlisting}
\end{revimprompt}

\begin{revimprompt}[suppgray]{Direct backbone reference}{supp:fig:direct}
\promptpart{Inputs}
\reviminput{video, timed text, question, options}.

\promptpart{Instruction}
Select the best answer to the multiple-choice question based on the supplied video and available timed text. Respond with the correct option letter in the specified JSON format.
\begin{lstlisting}[style=revimresponse]
{"choice":"B"}
\end{lstlisting}
\end{revimprompt}

\begin{revimprompt}[supppurple]{Referent alignment during encoding}{supp:fig:alignment}
\promptpart{Inputs}
\reviminput{observed argument and source account}, \reviminput{existing entity}, \reviminput{current target}.

\promptpart{Decide identity}
Match each referent by identity, using order or time to distinguish repeated events.
\promptpart{Response}
Return both judgments for each candidate. Use false for an unsupported match.
\begin{lstlisting}[style=revimresponse]
{
  "alignments": [{
    "same_existing_entity": false,
    "same_given_target": true}]}
\end{lstlisting}
\end{revimprompt}

\begin{revimprompt}[supporange]{Relation encoding and memory update}{supp:fig:encoding}
\promptpart{Inputs}
\reviminput{episode accounts and their sources}, \reviminput{supported integrations or corrections}, \reviminput{public relation forms}, \reviminput{read participants and current need}.

\promptpart{Index each account}
Reuse a schema when the predicate and role meanings match. Otherwise define typed roles and bind their observed values. Preserve the occurrence, support and temporal scope. Return bindings in account order.
\promptpart{Example binding fields}
\begin{lstlisting}[style=revimresponse]
{"predicate":"has_color","roles":[
  {"role":"entity","value":"rabbit"},
  {"role":"color","value":"blue"}],
 "source_ids":["F003"]}
\end{lstlisting}
\end{revimprompt}

\begin{revimprompt}[suppteal]{Video observation}{supp:fig:perception}
\promptpart{Acquired material}
\reviminput{executed request and focus}, \reviminput{question and options}, \reviminput{local need}, \reviminput{frames, sample times and aligned text}. For a media Read, the reopened episode account is also available.

\promptpart{Locate the occurrence}
Use the participant, event order and source time to identify the occurrence. Aligned speech can locate a scene. Inspect its frames for the requested attribute.
\promptpart{Describe what the return establishes}
Record the observed action, attribute, text or quantity with its source labels. Keep participants and occurrences distinct. If the target is unlocated, describe the observed scene and state what remains missing.
\promptpart{Write back the evidence}
Link each episode to the media inspected. Cite supporting records for a cross-episode conclusion. When new media corrects an earlier interpretation, identify the record and its replacement.
\promptpart{Example observation fields}
\begin{lstlisting}[style=revimresponse]
{
  "text": "The rabbit is blue.",
  "source_ids": ["F003"],
  "relations": []
}
\end{lstlisting}
\promptpart{Assess the local need}
Check the participant, requested attribute and scope against the return. A sighting alone leaves an attribute open.
\end{revimprompt}
\FloatBarrier

\onecolumn
\newenvironment{revimcasefigure}{\par\captionsetup{hypcap=false,type=figure}}{\par}
\section{Qualitative Analysis}
\label{supp:qualitative}
The qualitative examples use up to seven acquisition steps per question. Figures~\ref{supp:fig:family-start} and~\ref{supp:fig:board-diving} follow the same scoreboard across questions. Figure~\ref{supp:fig:experience-return} traces an observation procedure from the handshake question to the diving question.

\begin{revimcasefigure}
\begin{center}
\raisebox{-.5\height}{\includegraphics[trim=6bp 327bp 6bp 5bp,clip,width=.485\linewidth]{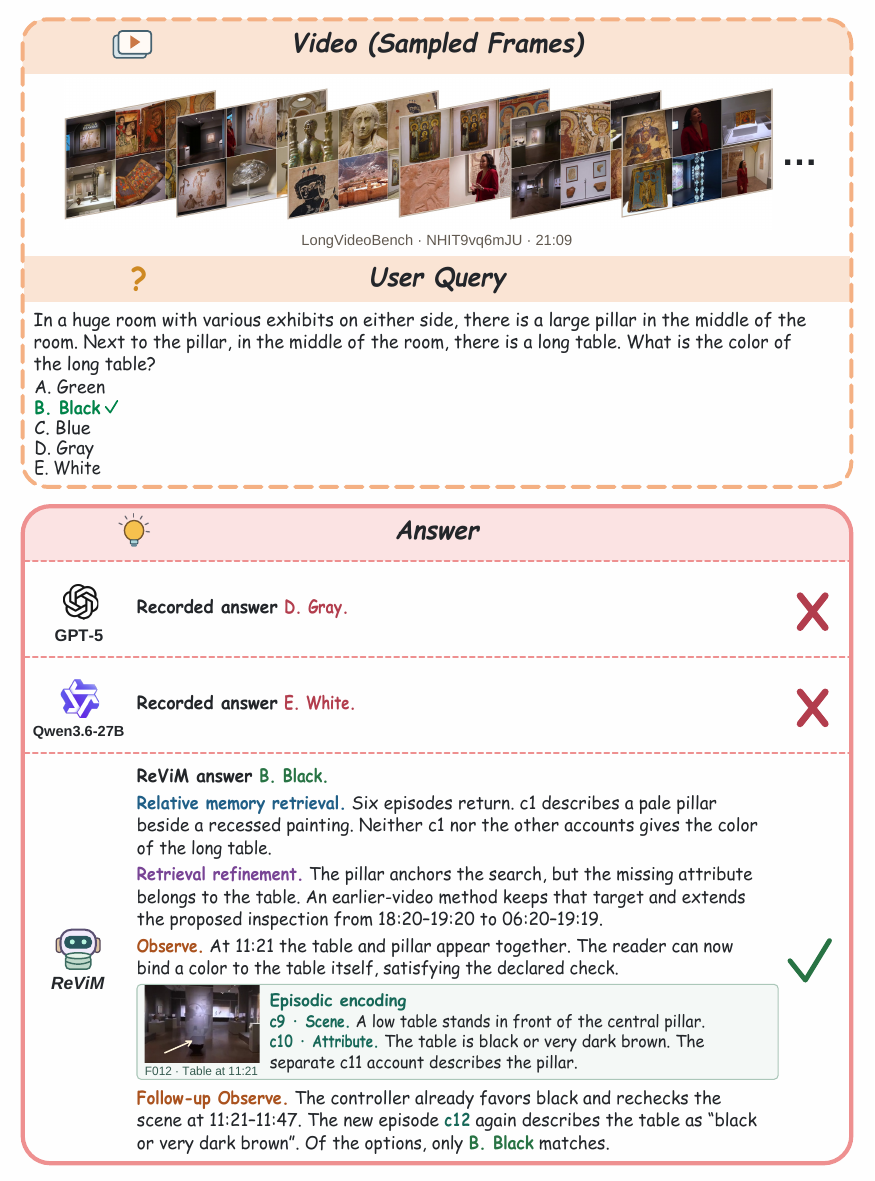}}\hfill
\raisebox{-.5\height}{\includegraphics[trim=6bp 5bp 6bp 239.9bp,clip,width=.495\linewidth]{01-GPT5-Qwen3.6-27B-ReViM-museum.pdf}}
\end{center}
\caption{\textbf{From a location to an attribute on LongVideoBench.} Search locates the pillar. An observation at 11:21 shows the table beside it, and a second inspection confirms a black or very dark brown surface. GPT-5 selects gray, Qwen3.6-27B selects white, and ReViM with Qwen3.6-27B selects black.}
\label{supp:fig:museum}
\end{revimcasefigure}

\smallskip\begin{revimcasefigure}
\noindent\makebox[\linewidth][c]{\includegraphics[width=\linewidth,height=.455\textheight,keepaspectratio]{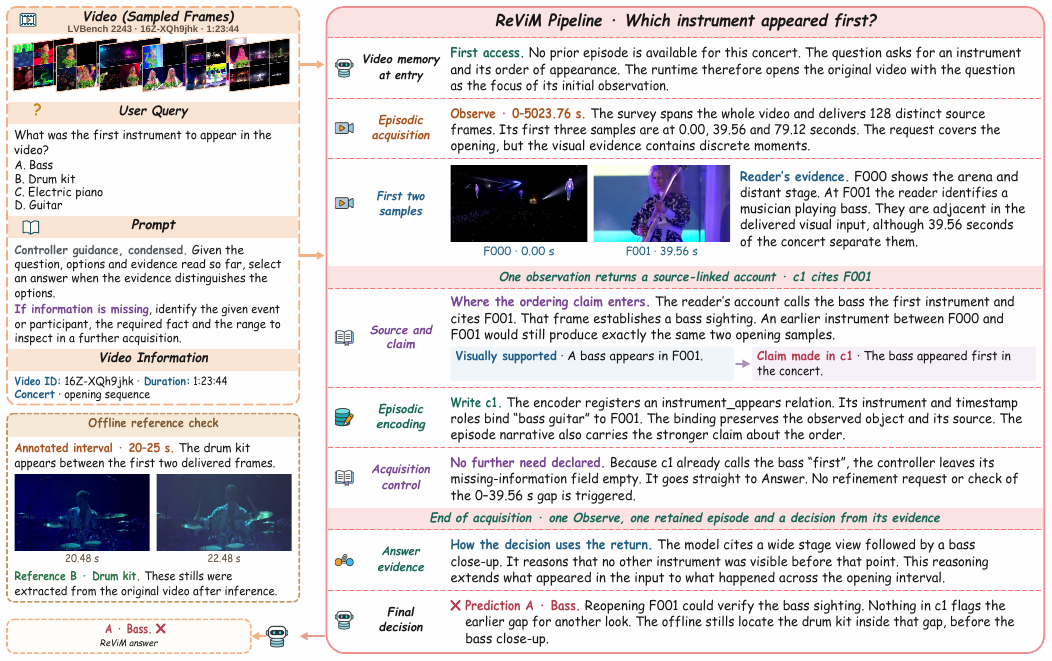}}\par
\caption{\textbf{A missed first occurrence on LVBench.} ReViM answers bass after one survey (39.56~s sighting) and submits no case. Post-evaluation frames show the drum kit at 20.48 and 22.48~s, between the first two samples.}
\label{supp:fig:first-instrument}
\end{revimcasefigure}

\newpage
\begin{revimcasefigure}
\noindent\makebox[\linewidth][c]{\includegraphics[trim=25bp 96bp 25bp 38bp,clip,width=\linewidth,height=.93\textheight,keepaspectratio]{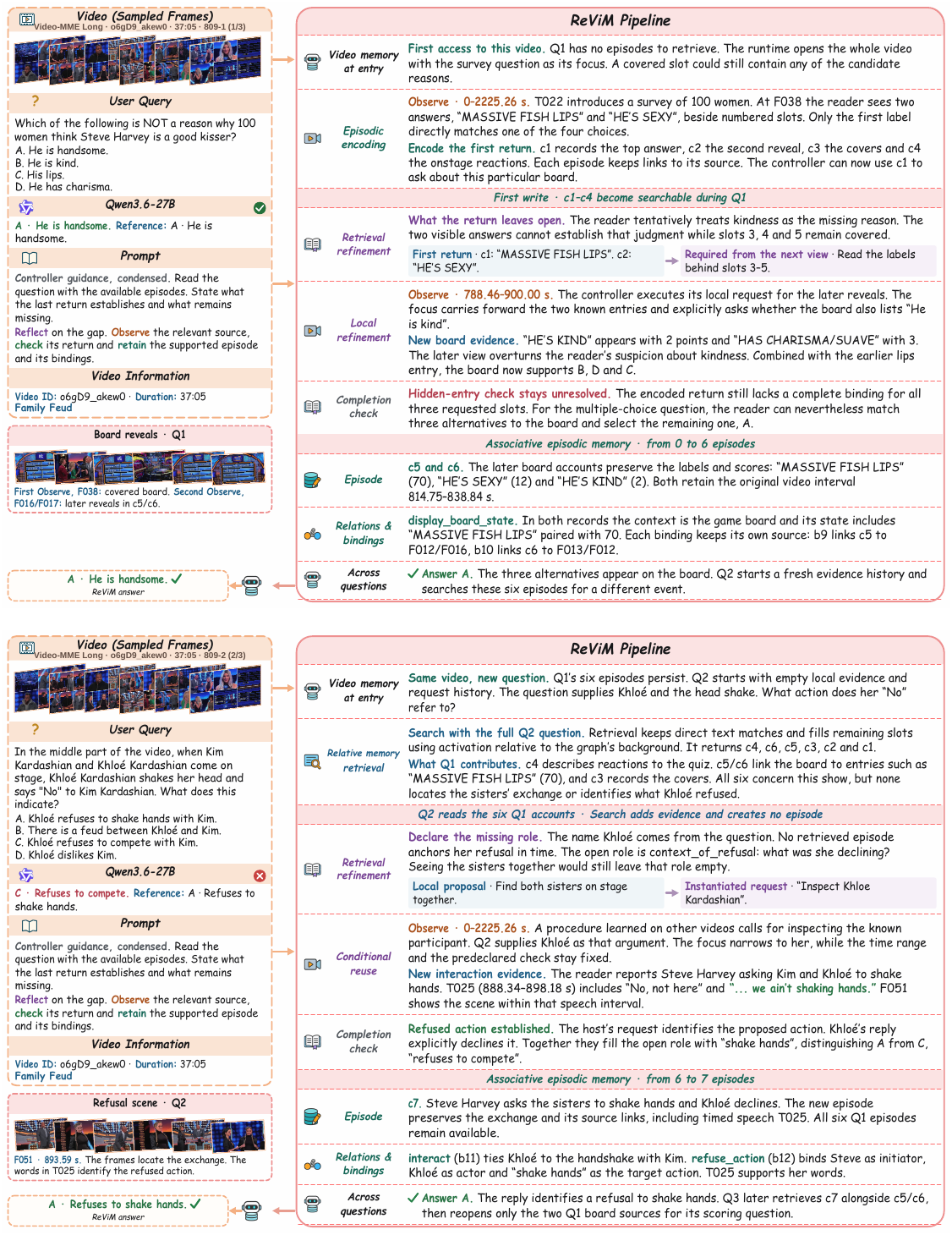}}\par
\caption{\textbf{Building and extending video memory on Video-MME Long.} Q1 starts with empty memory and retains six records, including the survey board. Q2 retrieves those records, but none identifies the action Khlo\'e refuses. The agent observes the exchange and adds a new episode linking her refusal to the host's request to shake hands. Both ReViM and Direct use Qwen3.6-27B.}
\label{supp:fig:family-start}
\end{revimcasefigure}

\newpage
\begin{revimcasefigure}
\noindent\makebox[\linewidth][c]{\includegraphics[trim=25bp 78bp 25bp 38bp,clip,width=\linewidth,height=.90\textheight,keepaspectratio]{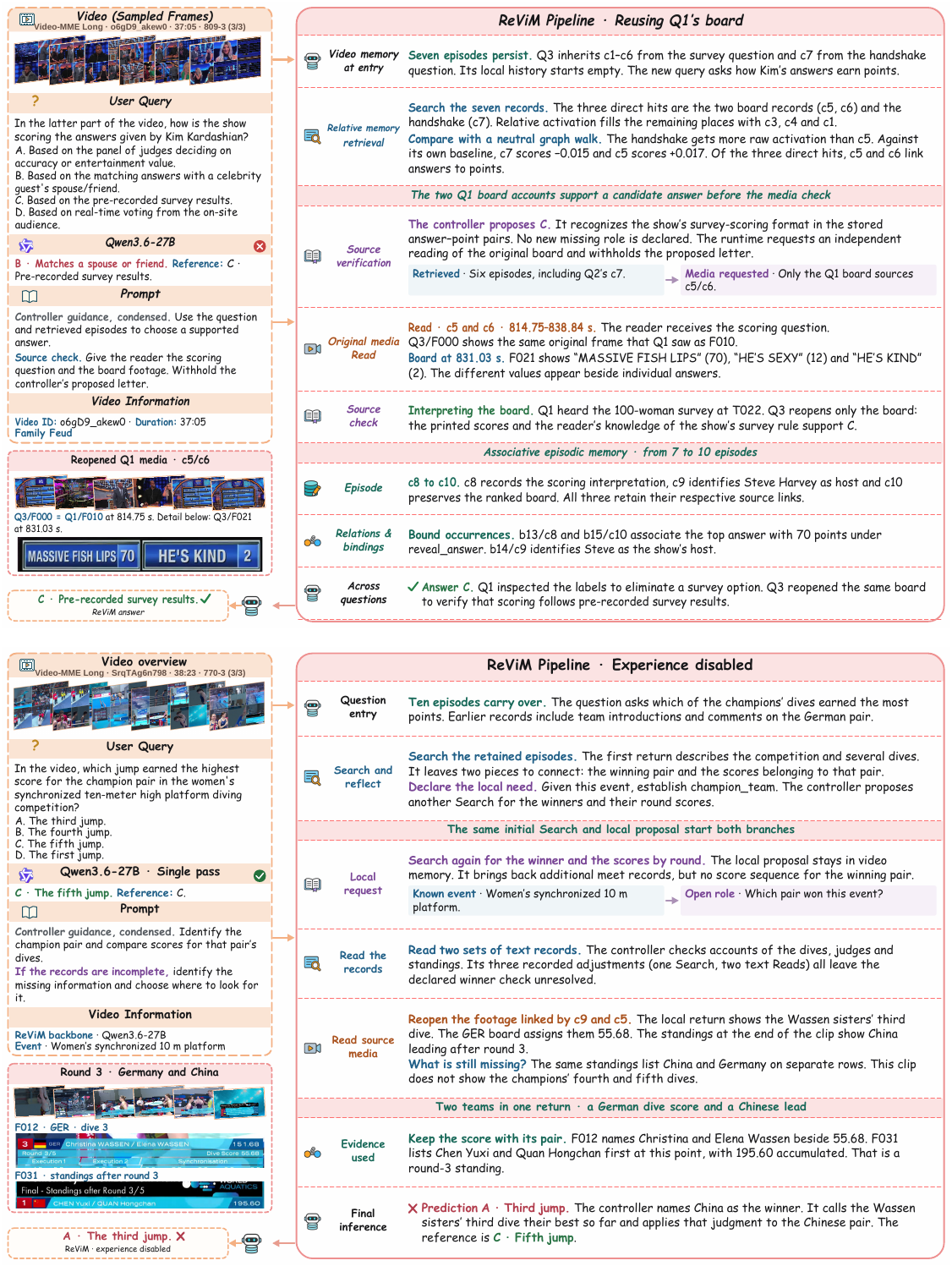}}\par
\caption{\textbf{Reading the same scoreboard across questions and distinguishing teams in a new video.} Top: Q3 reads point values from the board saved in Q1. Bottom: both 27B diving runs start from the same memory, decision and local request. With experience disabled, three adjustments remain unresolved. A media read shows Germany's third dive and China's lead after round three. The controller selects A (third dive) for China using evidence about Germany.}
\label{supp:fig:board-diving}
\end{revimcasefigure}

\newpage
\begin{revimcasefigure}
\noindent\makebox[\linewidth][c]{\includegraphics[trim=25bp 105bp 25bp 38bp,clip,width=\linewidth,height=.825\textheight,keepaspectratio]{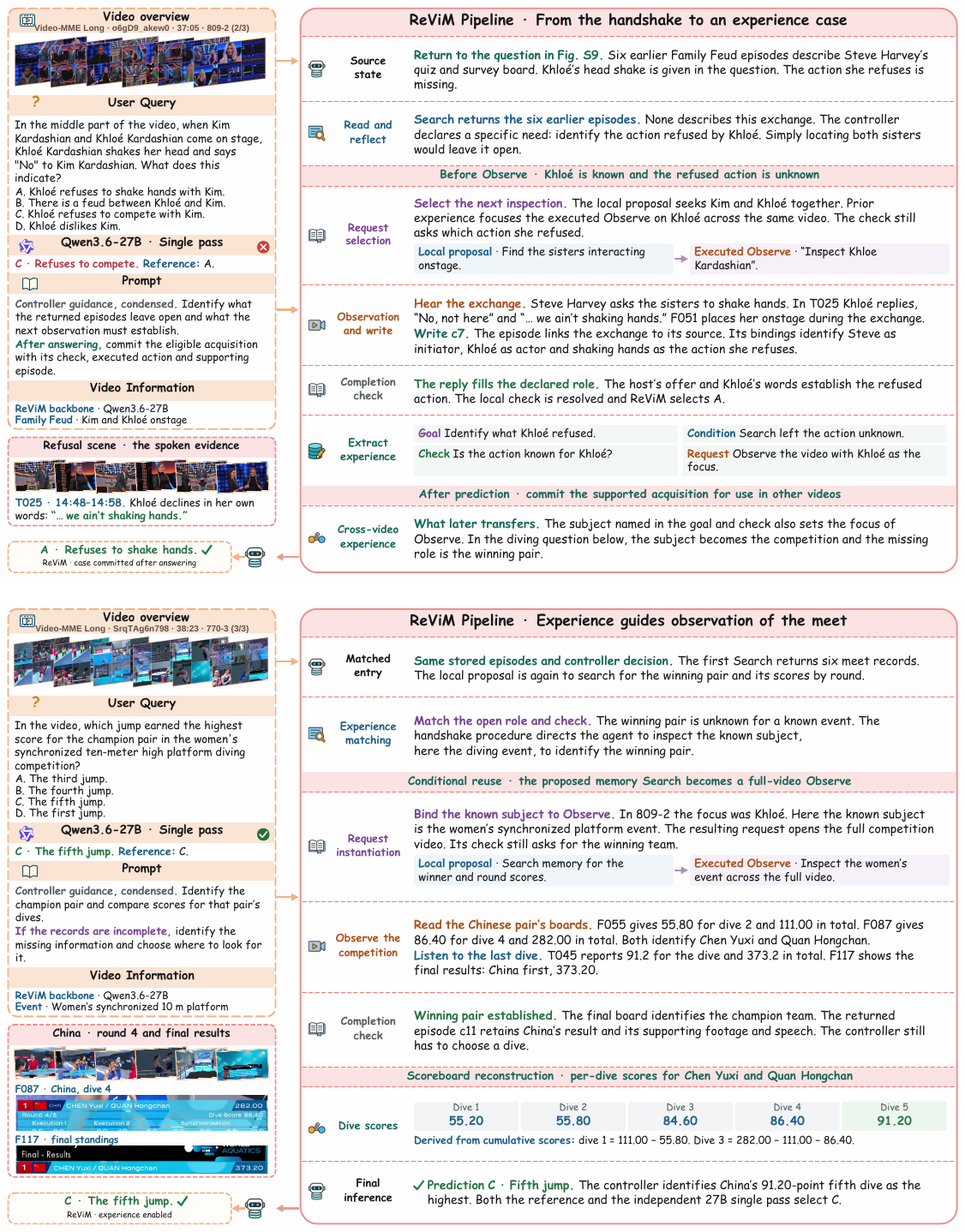}}\par
\caption{\textbf{From the handshake case to the diving question.} Top: Focusing on Khlo\'e reveals what she refused. After answering, ReViM retains the observation request together with its goal, condition and completion check. Bottom: ReViM applies the same procedure to the diving event, replacing a proposed memory search with an observation of the full competition. The final standings identify the Chinese pair as the champions. Their fifth dive earns their highest score (91.2), supporting C. The direct 27B baseline also selects C.}
\label{supp:fig:experience-return}
\end{revimcasefigure}

%% file: main.bib
@String(CVPR = {CVPR})

@String(ICCV = {ICCV})

@String(AAAI = {AAAI})

@inproceedings{wang2024videoagent,
  title={Videoagent: Long-form video understanding with large language model as agent},
  author={Wang, Xiaohan and Zhang, Yuhui and Zohar, Orr and Yeung-Levy, Serena},
  booktitle={European Conference on Computer Vision},
  pages={58--76},
  year={2024},
  organization={Springer},
}

@inproceedings{song2024moviechat,
  title={Moviechat: From dense token to sparse memory for long video understanding},
  author={Song, Enxin and Chai, Wenhao and Wang, Guanhong and Zhang, Yucheng and Zhou, Haoyang and Wu, Feiyang and Chi, Haozhe and Guo, Xun and Ye, Tian and Zhang, Yanting and others},
  booktitle={2024 IEEE/CVF Conference on Computer Vision and Pattern Recognition (CVPR)},
  pages={18221--18232},
  year={2024},
  organization={IEEE},
}

@article{liu2023visual,
  title={Visual instruction tuning},
  author={Liu, Haotian and Li, Chunyuan and Wu, Qingyang and Lee, Yong Jae},
  journal={Advances in neural information processing systems},
  volume={36},
  pages={34892--34916},
  year={2023},
}

@misc{bai2025qwen25vltechnicalreport,
  title={Qwen2.5-VL Technical Report},
  author={Shuai Bai and Keqin Chen and Xuejing Liu and Jialin Wang and Wenbin Ge and Sibo Song and Kai Dang and Peng Wang and Shijie Wang and Jun Tang and Humen Zhong and Yuanzhi Zhu and Mingkun Yang and Zhaohai Li and Jianqiang Wan and Pengfei Wang and Wei Ding and Zheren Fu and Yiheng Xu and Jiabo Ye and Xi Zhang and Tianbao Xie and Zesen Cheng and Hang Zhang and Zhibo Yang and Haiyang Xu and Junyang Lin},
  year={2025},
  eprint={2502.13923},
  archiveprefix={arXiv},
  primaryclass={cs.CV},
  url={https://arxiv.org/abs/2502.13923},
}

@article{chen2024expanding,
  title={Expanding performance boundaries of open-source multimodal models with model, data, and test-time scaling},
  author={Chen, Zhe and Wang, Weiyun and Cao, Yue and Liu, Yangzhou and Gao, Zhangwei and Cui, Erfei and Zhu, Jinguo and Ye, Shenglong and Tian, Hao and Liu, Zhaoyang and others},
  journal={arXiv preprint arXiv:2412.05271},
  year={2024},
}

@inproceedings{lin2024video,
  title={Video-llava: Learning united visual representation by alignment before projection},
  author={Lin, Bin and Ye, Yang and Zhu, Bin and Cui, Jiaxi and Ning, Munan and Jin, Peng and Yuan, Li},
  booktitle={Proceedings of the 2024 conference on empirical methods in natural language processing},
  pages={5971--5984},
  year={2024},
}

@article{cheng2024videollama,
  title={Videollama 2: Advancing spatial-temporal modeling and audio understanding in video-llms},
  author={Cheng, Zesen and Leng, Sicong and Zhang, Hang and Xin, Yifei and Li, Xin and Chen, Guanzheng and Zhu, Yongxin and Zhang, Wenqi and Luo, Ziyang and Zhao, Deli and others},
  journal={arXiv preprint arXiv:2406.07476},
  year={2024},
}

@article{zhang2024llava,
  title={Llava-video: Video instruction tuning with synthetic data},
  author={Zhang, Yuanhan and Wu, Jinming and Li, Wei and Li, Bo and Ma, Zejun and Liu, Ziwei and Li, Chunyuan},
  journal={arXiv preprint arXiv:2410.02713},
  year={2024},
}

@article{bai2025qwen3,
  title={Qwen3-vl technical report},
  author={Bai, Shuai and Cai, Yuxuan and Chen, Ruizhe and Chen, Keqin and Chen, Xionghui and Cheng, Zesen and Deng, Lianghao and Ding, Wei and Gao, Chang and Ge, Chunjiang and others},
  journal={arXiv preprint arXiv:2511.21631},
  year={2025},
}

@article{hong2025glm,
  title={{GLM-4.5V} and {GLM-4.1V-Thinking}: Towards Versatile Multimodal Reasoning with Scalable Reinforcement Learning},
  author={Hong, Wenyi and Yu, Wenmeng and Gu, Xiaotao and Wang, Guo and Gan, Guobing and Tang, Haomiao and Cheng, Jiale and Qi, Ji and Ji, Junhui and Pan, Lihang and others},
  journal={arXiv preprint arXiv:2507.01006},
  year={2025},
}

@article{zhang2024long,
  title={Long context transfer from language to vision},
  author={Zhang, Peiyuan and Zhang, Kaichen and Li, Bo and Zeng, Guangtao and Yang, Jingkang and Zhang, Yuanhan and Wang, Ziyue and Tan, Haoran and Li, Chunyuan and Liu, Ziwei},
  journal={arXiv preprint arXiv:2406.16852},
  year={2024},
}

@inproceedings{chen2025longvila,
  title={Longvila: Scaling long-context visual language models for long videos},
  author={Chen, Yukang and Xue, Fuzhao and Li, Dacheng and Hu, Qinghao and Zhu, Ligeng and Li, Xiuyu and Fang, Yunhao and Tang, Haotian and Yang, Shang and Liu, Zhijian and others},
  booktitle={International Conference on Learning Representations},
  volume={2025},
  pages={18227--18246},
  year={2025},
}

@article{shen2024longvu,
  title={Longvu: Spatiotemporal adaptive compression for long video-language understanding},
  author={Shen, Xiaoqian and Xiong, Yunyang and Zhao, Changsheng and Wu, Lemeng and Chen, Jun and Zhu, Chenchen and Liu, Zechun and Xiao, Fanyi and Varadarajan, Balakrishnan and Bordes, Florian and others},
  journal={arXiv preprint arXiv:2410.17434},
  year={2024},
}

@inproceedings{li2026videochat,
  title={Videochat-flash: Hierarchical compression for long-context video modeling},
  author={Li, Xinhao and Wang, Yi and Yu, Jiashuo and Zeng, Xiangyu and Zhu, Yuhan and Huang, Haian and Gao, Jianfei and Li, Kunchang and He, Yinan and Wang, Chenting and others},
  booktitle={International Conference on Learning Representations},
  volume={2026},
  pages={109089--109117},
  year={2026},
}

@article{wang2024retake,
  title={Retake: Reducing temporal and knowledge redundancy for long video understanding},
  author={Wang, Xiao and Si, Qingyi and Wu, Jianlong and Zhu, Shiyu and Cao, Li and Nie, Liqiang},
  journal={arXiv preprint arXiv:2412.20504},
  year={2024},
}

@inproceedings{wang2025adaretake,
  title={Adaretake: Adaptive redundancy reduction to perceive longer for video-language understanding},
  author={Wang, Xiao and Si, Qingyi and Zhu, Shiyu and Wu, Jianlong and Cao, Li and Nie, Liqiang},
  booktitle={Findings of the Association for Computational Linguistics: ACL 2025},
  pages={5417--5432},
  year={2025},
}

@inproceedings{tang2025adaptive,
  title={Adaptive keyframe sampling for long video understanding},
  author={Tang, Xi and Qiu, Jihao and Xie, Lingxi and Tian, Yunjie and Jiao, Jianbin and Ye, Qixiang},
  booktitle={2025 IEEE/CVF Conference on Computer Vision and Pattern Recognition (CVPR)},
  pages={29118--29128},
  year={2025},
  organization={IEEE},
}

@inproceedings{zhang2024simple,
  title={A simple llm framework for long-range video question-answering},
  author={Zhang, Ce and Lu, Taixi and Islam, Md Mohaiminul and Wang, Ziyang and Yu, Shoubin and Bansal, Mohit and Bertasius, Gedas},
  booktitle={Proceedings of the 2024 Conference on Empirical Methods in Natural Language Processing},
  pages={21715--21737},
  year={2024},
}

@inproceedings{luo2026video,
  author={Luo, Yongdong and Zheng, Xiawu and Li, Guilin and Yin, Shukang and Lin, Haojia and Fu, Chaoyou and Huang, Jinfa and Ji, Jiayi and Chao, Fei and Luo, Jiebo and Ji, Rongrong},
  booktitle={Advances in Neural Information Processing Systems},
  doi={10.52202/085713-5602},
  editor={D. Belgrave and C. Zhang and H. Lin and R. Pascanu and P. Koniusz and M. Ghassemi and N. Chen},
  pages={168008--168033},
  publisher={Curran Associates, Inc.},
  title={Video-RAG: Visually-aligned Retrieval-Augmented Long Video Comprehension},
  url={https://proceedings.neurips.cc/paper_files/paper/2025/file/f593c9c251d4d7cf14d4ab9861dfb7eb-Paper-Conference.pdf},
  volume={38, Main Conference},
  year={2025},
}

@inproceedings{jeong2025videorag,
  title={Videorag: Retrieval-augmented generation over video corpus},
  author={Jeong, Soyeong and Kim, Kangsan and Baek, Jinheon and Hwang, Sung Ju},
  booktitle={Findings of the Association for Computational Linguistics: ACL 2025},
  pages={21278--21298},
  year={2025},
}

@article{xue2025adavideorag,
  title={Adavideorag: Omni-contextual adaptive retrieval-augmented efficient long video understanding},
  author={Xue, Zhucun and Zhang, Jiangning and Xie, Xurong and Cai, Yuxuan and Liu, Yong and Li, Xiangtai and Tao, Dacheng},
  journal={arXiv preprint arXiv:2506.13589},
  year={2025},
}

@inproceedings{long2026seeing,
  title={Seeing, listening, remembering, and reasoning: A multimodal agent with long-term memory},
  author={Long, Lin and He, Yichen and Ye, Wentao and Pan, Yiyuan and Lin, Yuan and Li, Hang and Zhao, Junbo and Li, Wei},
  booktitle={International Conference on Learning Representations},
  volume={2026},
  pages={146197--146246},
  year={2026},
}

@inproceedings{yeo2026worldmm,
  title={Worldmm: Dynamic multimodal memory agent for long video reasoning},
  author={Yeo, Woongyeong and Kim, Kangsan and Yoon, Jaehong and Hwang, Sung Ju},
  booktitle={Proceedings of the IEEE/CVF Conference on Computer Vision and Pattern Recognition},
  pages={25599--25609},
  year={2026},
}

@inproceedings{choi2026keep,
  title={Keep it simple: Multi-key episodic memory retrieval for ultra-long video understanding},
  author={Choi, Yeeun and Yoo, Youngbeom and Lee, Joon-Young and Kang, Hyolim and Kim, Seon Joo},
  booktitle={European Conference on Computer Vision},
  pages={501--519},
  year={2026},
  organization={Springer},
}

@article{chen2026memdreamer,
  title={MemDreamer: Decoupling Perception and Reasoning for Long Video Understanding via Hierarchical Graph Memory and Agentic Retrieval Mechanism},
  author={Chen, Cong and Gan, Guo and Ji, Kaixiang and Zhang, ZhaoYang and Yang, Zhen and Yao, Guangming and Chen, Hao and Chen, Jingdong and Yuan, Yi and Shen, Chunhua},
  journal={arXiv preprint arXiv:2606.07512},
  year={2026},
}

@article{li2026lenswalk,
  title={LensWalk: Agentic video understanding by planning how you see in videos},
  author={Li, Keliang and Li, Yansong and Shen, Hongze and Liu, Mengdi and Chang, Hong and Shan, Shiguang},
  journal={arXiv preprint arXiv:2603.24558},
  year={2026},
}

@article{lin2026videoseek,
  title={VideoSeek: Long-Horizon Video Agent with Tool-Guided Seeking},
  author={Lin, Jingyang and Wu, Jialian and Liu, Jiang and Sun, Ximeng and Wang, Ze and Yu, Xiaodong and Luo, Jiebo and Liu, Zicheng and Barsoum, Emad},
  journal={arXiv preprint arXiv:2603.20185},
  year={2026},
}

@inproceedings{zhao2024expel,
  title={Expel: Llm agents are experiential learners},
  author={Zhao, Andrew and Huang, Daniel and Xu, Quentin and Lin, Matthieu and Liu, Yong-Jin and Huang, Gao},
  booktitle={Proceedings of the AAAI Conference on Artificial Intelligence},
  volume={38},
  number={17},
  pages={19632--19642},
  year={2024},
}

@article{wang2024agent,
  title={Agent workflow memory},
  author={Wang, Zora Zhiruo and Mao, Jiayuan and Fried, Daniel and Neubig, Graham},
  journal={arXiv preprint arXiv:2409.07429},
  year={2024},
}

@article{guo2025seed1,
  title={{Seed1.5-VL} Technical Report},
  author={Guo, Dong and Wu, Faming and Zhu, Feida and Leng, Fuxing and Shi, Guang and Chen, Haobin and Fan, Haoqi and Wang, Jian and Jiang, Jianyu and Wang, Jiawei and others},
  journal={arXiv preprint arXiv:2505.07062},
  year={2025},
}

@article{yang2023mm,
  title={Mm-react: Prompting chatgpt for multimodal reasoning and action},
  author={Yang, Zhengyuan and Li, Linjie and Wang, Jianfeng and Lin, Kevin and Azarnasab, Ehsan and Ahmed, Faisal and Liu, Zicheng and Liu, Ce and Zeng, Michael and Wang, Lijuan},
  journal={arXiv preprint arXiv:2303.11381},
  year={2023},
}

@inproceedings{suris2023vipergpt,
  title={Vipergpt: Visual inference via python execution for reasoning},
  author={Sur{\'\i}s, D{\'\i}dac and Menon, Sachit and Vondrick, Carl},
  booktitle={2023 IEEE/CVF International Conference on Computer Vision (ICCV)},
  pages={11854--11864},
  year={2023},
  organization={IEEE},
}

@article{yao2022react,
  title={React: Synergizing reasoning and acting in language models},
  author={Yao, Shunyu and Zhao, Jeffrey and Yu, Dian and Du, Nan and Shafran, Izhak and Narasimhan, Karthik and Cao, Yuan},
  journal={arXiv preprint arXiv:2210.03629},
  year={2022},
}

@inproceedings{wang2025videotree,
  title={Videotree: Adaptive tree-based video representation for llm reasoning on long videos},
  author={Wang, Ziyang and Yu, Shoubin and Stengel-Eskin, Elias and Yoon, Jaehong and Cheng, Feng and Bertasius, Gedas and Bansal, Mohit},
  booktitle={2025 IEEE/CVF Conference on Computer Vision and Pattern Recognition (CVPR)},
  pages={3272--3282},
  year={2025},
  organization={IEEE},
}

@inproceedings{ma2025drvideo,
  title={Drvideo: Document retrieval based long video understanding},
  author={Ma, Ziyu and Gou, Chenhui and Shi, Hengcan and Sun, Bin and Li, Shutao and Rezatofighi, Hamid and Cai, Jianfei},
  booktitle={2025 IEEE/CVF Conference on Computer Vision and Pattern Recognition (CVPR)},
  pages={18936--18946},
  year={2025},
  organization={IEEE},
}

@inproceedings{zhang2026deep,
  author={Zhang, Xiaoyi and Jia, Zhaoyang and Guo, Zongyu and Li, Jiahao and Li, Bin and Li, Houqiang and Lu, Yan},
  booktitle={Advances in Neural Information Processing Systems},
  doi={10.52202/085713-3005},
  editor={D. Belgrave and C. Zhang and H. Lin and R. Pascanu and P. Koniusz and M. Ghassemi and N. Chen},
  pages={89863--89895},
  publisher={Curran Associates, Inc.},
  title={Deep Video Discovery: Agentic Search with Tool Use for Long-form Video Understanding},
  url={https://proceedings.neurips.cc/paper_files/paper/2025/file/8190b210e9808e54ee16263b673a847d-Paper-Conference.pdf},
  volume={38, Main Conference},
  year={2025},
}

@inproceedings{he2024ma,
  title={Ma-lmm: Memory-augmented large multimodal model for long-term video understanding},
  author={He, Bo and Li, Hengduo and Jang, Young Kyun and Jia, Menglin and Cao, Xuefei and Shah, Ashish and Shrivastava, Abhinav and Lim, Ser-Nam},
  booktitle={2024 IEEE/CVF Conference on Computer Vision and Pattern Recognition (CVPR)},
  pages={13504--13514},
  year={2024},
  organization={IEEE},
}

@article{lewis2020retrieval,
  title={Retrieval-augmented generation for knowledge-intensive nlp tasks},
  author={Lewis, Patrick and Perez, Ethan and Piktus, Aleksandra and Petroni, Fabio and Karpukhin, Vladimir and Goyal, Naman and K{\"u}ttler, Heinrich and Lewis, Mike and Yih, Wen-tau and Rockt{\"a}schel, Tim and others},
  journal={Advances in neural information processing systems},
  volume={33},
  pages={9459--9474},
  year={2020},
}

@inproceedings{arefeen2024irag,
  title={irag: Advancing rag for videos with an incremental approach},
  author={Arefeen, Md Adnan and Debnath, Biplob and Uddin, Md Yusuf Sarwar and Chakradhar, Srimat},
  booktitle={Proceedings of the 33rd ACM International Conference on Information and Knowledge Management},
  pages={4341--4348},
  year={2024},
}

@inproceedings{lian2026verbatim,
  title={From verbatim to gist: Distilling pyramidal multimodal memory via semantic information bottleneck for long-horizon video agents},
  author={Lian, Niu and Wang, Yuting and Yao, Hanshu and Wang, Jinpeng and Chen, Bin and Wang, Yaowei and Zhang, Min and Xia, Shu-Tao},
  booktitle={Proceedings of the 64th Annual Meeting of the Association for Computational Linguistics (Volume 1: Long Papers)},
  pages={11601--11617},
  year={2026},
}

@inproceedings{zuo2026videolucy,
  author={Zuo, Jialong and Deng, Yongtai and Kong, Lingdong and Yang, Jingkang and Jin, Rui and Zhang, Yiwei and Sang, Nong and Pan, Liang and Liu, Ziwei and Gao, Changxin},
  booktitle={Advances in Neural Information Processing Systems},
  doi={10.52202/085713-0864},
  editor={D. Belgrave and C. Zhang and H. Lin and R. Pascanu and P. Koniusz and M. Ghassemi and N. Chen},
  pages={25660--25691},
  publisher={Curran Associates, Inc.},
  title={VideoLucy: Deep Memory Backtracking for Long Video Understanding},
  url={https://proceedings.neurips.cc/paper_files/paper/2025/file/24a2b1e2d4c3b20f09eb006d3d610777-Paper-Conference.pdf},
  volume={38, Main Conference},
  year={2025},
}

@misc{chen2026sproutbuildingdynamicmemory,
  title={Sprout: Building Dynamic Memory While Reasoning for Agentic Video Understanding},
  author={Wei Chen and Xuanyu Zheng and Yancheng Long and Haoyang Xu and Kaiyu Jiang and Bin Wen and Tingting Gao and Han Li and Long Chen},
  year={2026},
  eprint={2609.35497},
  archiveprefix={arXiv},
  primaryclass={cs.CV},
  url={https://arxiv.org/abs/2609.35497},
}

@misc{liu2026videotapestryqueryadaptivememoryrefinement,
  title={VideoTapestry: Query-Adaptive Memory Refinement for Multi-Agent Long-Video Understanding},
  author={Yucheng Liu and Yufei Yin and Mingxiao Feng and Jiajun Deng and Wengang Zhou and Houqiang Li},
  year={2026},
  eprint={2610.06672},
  archiveprefix={arXiv},
  primaryclass={cs.CV},
  url={https://arxiv.org/abs/2610.06672},
}

@article{gutierrez2024hipporag,
  title={Hipporag: Neurobiologically inspired long-term memory for large language models},
  author={Guti{\'e}rrez, Bernal J and Shu, Yiheng and Gu, Yu and Yasunaga, Michihiro and Su, Yu},
  journal={Advances in neural information processing systems},
  volume={37},
  pages={59532--59569},
  year={2024},
}

@inproceedings{wang2025lvbench,
  title={Lvbench: An extreme long video understanding benchmark},
  author={Wang, Weihan and He, Zehai and Hong, Wenyi and Cheng, Yean and Zhang, Xiaohan and Qi, Ji and Ding, Ming and Gu, Xiaotao and Huang, Shiyu and Xu, Bin and others},
  booktitle={2025 IEEE/CVF International Conference on Computer Vision (ICCV)},
  pages={22958--22967},
  year={2025},
  organization={IEEE},
}

@article{wu2024longvideobench,
  title={Longvideobench: A benchmark for long-context interleaved video-language understanding},
  author={Wu, Haoning and Li, Dongxu and Chen, Bei and Li, Junnan},
  journal={Advances in Neural Information Processing Systems},
  volume={37},
  pages={28828--28857},
  year={2024},
}

@inproceedings{fu2025video,
  title={Video-mme: The first-ever comprehensive evaluation benchmark of multi-modal llms in video analysis},
  author={Fu, Chaoyou and Dai, Yuhan and Luo, Yongdong and Li, Lei and Ren, Shuhuai and Zhang, Renrui and Wang, Zihan and Zhou, Chenyu and Shen, Yunhang and Zhang, Mengdan and others},
  booktitle={2025 IEEE/CVF Conference on Computer Vision and Pattern Recognition (CVPR)},
  pages={24108--24118},
  year={2025},
  organization={IEEE},
}

@article{comanici2025gemini,
  title={Gemini 2.5: Pushing the frontier with advanced reasoning, multimodality, long context, and next generation agentic capabilities},
  author={Comanici, Gheorghe and Bieber, Eric and Schaekermann, Mike and Pasupat, Ice and Sachdeva, Noveen and Dhillon, Inderjit and Blistein, Marcel and Ram, Ori and Zhang, Dan and Rosen, Evan and others},
  journal={arXiv preprint arXiv:2507.06261},
  year={2025},
}

@article{hurst2024gpt,
  title={Gpt-4o system card},
  author={Hurst, Aaron and Lerer, Adam and Goucher, Adam P and Perelman, Adam and Ramesh, Aditya and Clark, Aidan and Ostrow, AJ and Welihinda, Akila and Hayes, Alan and Radford, Alec and others},
  journal={arXiv preprint arXiv:2410.21276},
  year={2024},
}

@inproceedings{mangalam2023egoschema,
  title={EgoSchema: A Diagnostic Benchmark for Very Long-form Video Language Understanding},
  author={Mangalam, Karttikeya and Akshulakov, Raiymbek and Malik, Jitendra},
  booktitle={Advances in Neural Information Processing Systems},
  editor={Oh, A. and Naumann, T. and Globerson, A. and Saenko, K. and Hardt, M. and Levine, S.},
  pages={46212--46244},
  publisher={Curran Associates, Inc.},
  doi={10.52202/075280-2004},
  volume={36},
  year={2023},
  url={https://proceedings.neurips.cc/paper_files/paper/2023/hash/90ce332aff156b910b002ce4e6880dec-Abstract-Datasets_and_Benchmarks.html}
}

@misc{openai2025gpt5systemcard,
  title={{GPT-5} System Card},
  author={{OpenAI}},
  year={2025},
  url={https://openai.com/index/gpt-5-system-card/}
}

@misc{qwen2026qwen36modelcard,
  title={{Qwen3.6-27B} Model Card},
  author={{Qwen Team}},
  year={2026},
  url={https://huggingface.co/Qwen/Qwen3.6-27B}
}

@inproceedings{chen-etal-2024-m3,
  title={{M3}-Embedding: Multi-Linguality, Multi-Functionality, Multi-Granularity Text Embeddings Through Self-Knowledge Distillation},
  author={Chen, Jianlyu and Xiao, Shitao and Zhang, Peitian and Luo, Kun and Lian, Defu and Liu, Zheng},
  booktitle={Findings of the Association for Computational Linguistics: ACL 2024},
  year={2024},
  pages={2318--2335},
  publisher={Association for Computational Linguistics},
  doi={10.18653/v1/2024.findings-acl.137},
  url={https://aclanthology.org/2024.findings-acl.137/}
}

@misc{pang2025mrvideo,
  title={{MR. Video}: {MapReduce} is the Principle for Long Video Understanding},
  author={Pang, Ziqi and Wang, Yu-Xiong},
  year={2025},
  eprint={2504.16082},
  archivePrefix={arXiv},
  primaryClass={cs.CV},
  url={https://arxiv.org/abs/2504.16082}
}

@inproceedings{wang2026active,
  title={Active video perception: Iterative evidence seeking for agentic long video understanding},
  author={Wang, Ziyang and Zhou, Honglu and Wang, Shijie and Li, Junnan and Xiong, Caiming and Savarese, Silvio and Bansal, Mohit and Ryoo, Michael S and Niebles, Juan Carlos},
  booktitle={Proceedings of the IEEE/CVF Conference on Computer Vision and Pattern Recognition},
  pages={9088--9099},
  year={2026},
}

@inproceedings{pmlr-v202-radford23a,
  title={Robust Speech Recognition via Large-Scale Weak Supervision},
  author={Radford, Alec and Kim, Jong Wook and Xu, Tao and Brockman, Greg and Mcleavey, Christine and Sutskever, Ilya},
  booktitle={Proceedings of the 40th International Conference on Machine Learning},
  pages={28492--28518},
  year={2023},
  volume={202},
  series={Proceedings of Machine Learning Research},
  publisher={PMLR},
  url={https://proceedings.mlr.press/v202/radford23a.html}
}

@inproceedings{yin2026videoarm,
  title={Videoarm: Agentic reasoning over hierarchical memory for long-form video understanding},
  author={Yin, Yufei and Meng, Qianke and Chen, Minghao and Ding, Jiajun and Shao, Zhenwei and Yu, Zhou},
  booktitle={Proceedings of the IEEE/CVF Conference on Computer Vision and Pattern Recognition},
  pages={24042--24051},
  year={2026}
}

@inproceedings{wang2026routing,
  title={Routing before looking: Query-adaptive evidence acquisition for long-form video understanding},
  author={Wang, Tianyue and Wu, Xuying and Ma, Yuxiang and Liang, Ruiming and Kang, Jiaxuan and Hao, Yanchao and Wei, Zheng and Qu, Leigang and Guo, Haiyun and Wang, Jinqiao},
  booktitle={Findings of the Association for Computational Linguistics: EMNLP 2026},
  pages={42290--42308},
  year={2026}
}


%% file: supplement_refs.bib
@inproceedings{huang2026meta,
  author    = {Huang, Jing and Chen, Luyuan and Xu, Zhijie and Li, Yadong and Xu, Xingzhong and Chen, Siye and Liu, Jie and Kong, Ming and Zhu, Qiang},
  title     = {{META}: Meta Evolution of Tool Trajectory Adaptation for Long-Video Understanding},
  booktitle = {Proceedings of the IEEE/CVF Conference on Computer Vision and Pattern Recognition (CVPR)},
  month     = {June},
  year      = {2026},
  pages     = {9837--9846}
}
